\documentclass[journal]{IEEEtran}
\usepackage{graphicx}
\usepackage{float}
\usepackage{makecell}
\usepackage{array}
\usepackage{xcolor}
\usepackage{amsmath}
\usepackage{mathtools}
\usepackage{colortbl}
\usepackage{booktabs}
\usepackage{algorithm}
\usepackage{algorithmic}
\usepackage{diagbox}
\usepackage{multirow}
\usepackage{subfigure}
\usepackage{tabularx}
\usepackage{amssymb}
\usepackage{pifont}
\usepackage{multicol}
\usepackage{marvosym}
\usepackage{placeins}
\usepackage{stfloats}
\usepackage{needspace}
\usepackage{adjustbox}
\usepackage{enumitem}
\usepackage[activate={true,nocompatibility},final,tracking=true,kerning=true,spacing=true,factor=1100,stretch=10,shrink=10]{microtype}
\usepackage{hyperref}
\newcommand{\rev}[1]{#1}
\newenvironment{revised}{}{}
\def\BibTeX{{\rm B\kern-.05em{\sc i\kern-.025em b}\kern-.08em
    T\kern-.1667em\lower.7ex\hbox{E}\kern-.125emX}}

\makeatletter
\providecommand{\sf@counterlist}{}  
\@ifpackageloaded{caption}{\typeout{=== caption.sty has been loaded ===}}{}
\def\@IEEEtablecaptionseplist{\relax} 
\makeatother

\begin{document}

\title{C-TRAIL: A Commonsense World Framework for Trajectory Planning in Autonomous Driving}

\author{Zhihong Cui\textsuperscript{1},\, 
        Haoran Tang\textsuperscript{2},\, 
        Tianyi Li\textsuperscript{3},\, 
        Yushuai Li\textsuperscript{3},\, 
        Peiyuan Guan\textsuperscript{1\Letter},\, 
        Amir Taherkordi\textsuperscript{1},\, 
        Tor Skeie\textsuperscript{1}%
\thanks{\textsuperscript{1} Zhihong Cui, Peiyuan Guan, Amir Taherkordi, and Tor Skeie are with the Department of Informatics, University of Oslo, Norway. Email: \{zhihongc, peiyuang, amirhost, tskeie\}@ifi.uio.no}
\thanks{\textsuperscript{2} Haoran Tang is with the Department of Computing, Hong Kong Polytechnic University, Hong Kong, China. Email: haoran.tang@connect.polyu.hk}
\thanks{\textsuperscript{3} Tianyi Li and Yushuai Li are with the Department of Computer Science, Aalborg University, Denmark. Email: \{tianyi, yusli\}@cs.aau.dk}
\thanks{\textsuperscript{\Letter}Corresponding author: Peiyuan Guan}
}

\markboth{IEEE Transactions on Vehicular Technology}%
{Cui \MakeLowercase{\textit{et al.}}: C-TRAIL: A Commonsense World Framework for Trajectory Planning in Autonomous Driving}
\maketitle

\begin{abstract}
\begin{revised}
Trajectory planning for autonomous driving increasingly leverages large language models (LLMs) for commonsense reasoning, yet LLM outputs are inherently unreliable, posing risks in safety-critical applications. We propose C-TRAIL, a framework built on a \emph{Commonsense World} that couples LLM-derived commonsense with a trust mechanism to guide trajectory planning. C-TRAIL operates through a closed-loop Recall, Plan, and Update cycle: the Recall module queries an LLM for semantic relations and quantifies their reliability via a dual-trust mechanism; the Plan module injects trust-weighted commonsense into Monte Carlo Tree Search (MCTS) through a Dirichlet trust policy; and the Update module adaptively refines trust scores and policy parameters from environmental feedback. Experiments on four simulated scenarios in Highway-env and two real-world levelXData datasets (highD, rounD) show that C-TRAIL consistently outperforms state-of-the-art baselines, reducing ADE by 40.2\%, FDE by 51.7\%, and improving SR by \rev{16.9\,pp} on average. The source code is available at \url{https://github.com/ZhihongCui/CTRAIL}.
\end{revised}
\end{abstract}

\IEEEpeerreviewmaketitle
\section{Introduction}

\IEEEPARstart{T}{rajectory} planning \cite{hu2023planning,chen2024planning} in a dynamic environment is essential for autonomous driving. For instance, when instructed to “Drive to the nearest gas station,” a vehicle must analyze its surroundings, such as nearby vehicles, gas stations, and lane structures, while continuously planning future trajectories \cite{li2023trajectory}. However, given the unpredictability of the autonomous driving environment, vehicles may encounter previously unseen environments, making trajectory planning generalization in unknown scenarios a demanding challenge. 

A common method to address this challenge is \textbf{world models}, which build internal representations of the environment from past experiences to support generalization \cite{groves2014framework,xie2023cognition,zhang2024predicting}. However, such models require large amounts of similar data to fit unseen scenarios, making them prone to dataset bias, overfitting, and inaccurate scenario understanding \cite{codevilla2019exploring, teng2023motion}.

Drawing inspiration from a profound question~\cite{epstein2017cognitive,son2024replay}: "Why can humans make route planning in unfamiliar cities just by looking at a map?” These works explore the core principles that underlie human planning skills and raise a pivotal distinction: humans seamlessly recall relevant commonsense knowledge (e.g., understanding of roads, traffic rules, and landmarks) from memory, enabling them to mentally plan and evaluate trajectories before executing any actions in reality. This observation catalyzes a fundamental question: \textbf{How can we introduce such commonsense-driven imaginative capabilities into trajectory planning?} 

\begin{figure}[htp]
    \centering
    \includegraphics[width=0.94\linewidth]{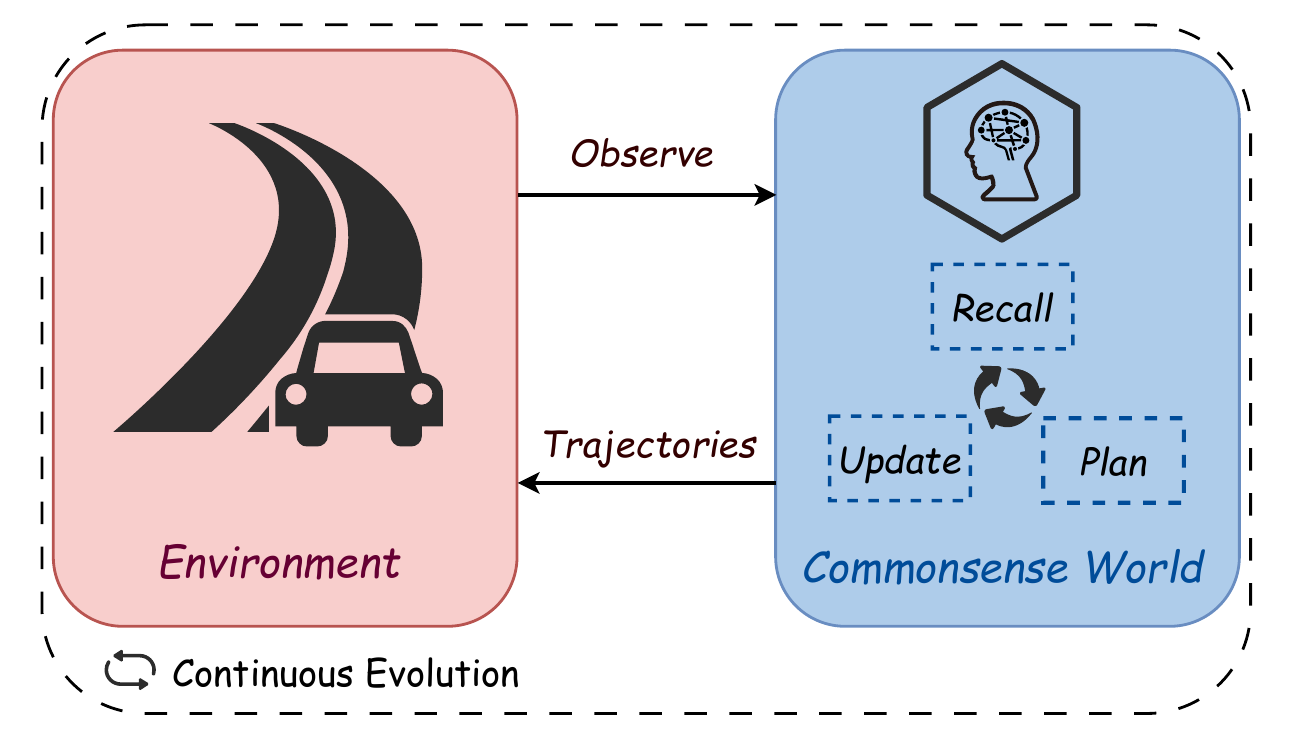}
    \caption{The commonsense world paradigm for trajectory planning in dynamic environments}
    \label{humanplanningparadigm}
\end{figure}

Recent advancements in LLMs \cite{song2024trial,felemban2024imotion,li2025synergy,tang2025dybooster} have demonstrated remarkable capabilities in commonsense reasoning and knowledge retrieval, positioning them as ideal repositories for enabling human-like trajectory planning. As a result, LLMs have been increasingly integrated into trajectory planning frameworks. 
Some methods \cite{peng2023check,huang2025survey} treat LLMs as external knowledge sources, querying them for contextual information or action suggestions based on past observations and environmental cues.
Others \cite{wang2024q,li2024embodied} emphasize multi-step action generation, leveraging LLMs to directly produce sequential trajectories.
A few recent efforts \cite{sun2023adaplanner,pan2023automatically} attempt to incorporate feedback-based refinement, enabling planning modules to adjust their strategies based on the output.
While these methods provide valuable insights, they are fundamentally fragmented and fall short of replicating the commonsense way that humans plan trajectories.
To bridge this critical gap, we propose a novel three-stage commonsense-driven trajectory planning paradigm: \textbf{Recall}, \textbf{Plan}, and \textbf{Update}.
As shown in Fig.~\ref{humanplanningparadigm}, this paradigm envisions a conceptual \textbf{Commonsense World} that interacts with the environment through a three-stage process. In the \textbf{Recall} stage, relevant knowledge is retrieved from LLMs to contextualize the current decision. In the \textbf{Plan} stage, multiple trajectories are simulated and evaluated. In the \textbf{Update} stage, the model is refined based on real-world feedback. To assess how existing methods relate to this paradigm, we analyze representative algorithms across these three stages and highlight their limitations.

In the \textbf{Recall} stage, existing methods \cite{mahmud2025integrating,zheng2023trafficsafetygpt,wang2023survey} use LLMs to retrieve domain-specific knowledge (e.g., traffic laws and route constraints) as contextual input for planning.
For instance, Ahn et al. \cite{brohan2023can} query an LLM to generate high-level action proposals for trajectory planning, thereby expanding the decision space with commonsense-driven directives. Zhao et al. \cite{arora2022ask} explore prompt engineering techniques to enhance the quality and relevance of retrieved knowledge.
Similarly, Huang et al. \cite{huang2022language} employ pre-trained LLMs to provide heuristic planning rules in simulated environments. However, our investigation reveals a key limitation.
Although LLMs are trained on massive datasets, their understanding of unseen scenarios is largely based on probabilistic pattern matching rather than genuine logical reasoning.
As a result, the recalled commonsense may be inconsistent with the actual environment, potentially leading to unsafe or suboptimal decisions.
Therefore, \textbf{assessing the trust of recalled knowledge before it is used in planning is paramount.}

In the \textbf{Plan} stage, LLMs act as planners \cite{silver2017mastering, guan2023leveraging, song2023llm} by directly generating the trajectory. However, planners driven solely by LLM \cite{song2023llm,liang2023code} remain constrained by the learned patterns inherent to their training data, substantially limiting their generalization capabilities in unseen scenarios. To address this, recent efforts have explored integrating LLMs with MCTS \cite{schrittwieser2020mastering, meta2022human}. In some methods \cite{liu2024evolution}, the LLM is treated as a high-level heuristic policy to guide the search. While promising, such integration often lacks adaptive mechanisms to regulate the LLM’s influence, leading to inconsistent guidance during planning.
Other methods \cite{sun2024prompt,du2023guiding} use LLMs to generate initial action priors (e.g., probability distributions over potential actions) to guide MCTS exploration. However, such priors are typically uncalibrated, making it difficult to determine how much influence they should exert on the search process. Thus, \textbf{ensuring that planning systems can selectively trust LLM guidance is a critical requirement.}

In the \textbf{Update} stage, real-time adaptability is essential for planning systems operating in dynamic environments.
Static LLM-based planners \cite{wang2023voyager, wu2023tidybot} risk becoming outdated and unreliable over time, leading to performance degradation. To address this, existing methods \cite{wang2024large, wu2024dlora, tang2025model} have explored various dynamic update strategies. One common method \cite{han2023ieta} is incremental fine-tuning, which continuously updates the LLM using new data collected during deployment. While this enables long-term adaptation, it typically requires large amounts of feedback data and high computational costs.
Another line of work focuses on policy-level updates \cite{zhang2024agent, tian2023learning, li2025optimal}, adjusting the planning policy based on recent observations or rewards.
Although more lightweight, these methods often fail to understand context-aware scenarios, limiting their robustness in complex scenarios. \rev{Moreover, existing update methods adjust the planning policy based on raw feedback signals but do not explicitly track or update the reliability of the knowledge source itself, making it impossible to distinguish failures caused by unreliable commonsense from those caused by environmental stochasticity.} \textbf{\rev{This highlights the need for a lightweight update mechanism that adaptively refines trust in commonsense knowledge over time.}}
 
Following the above analysis, we propose C-TRAIL, a commonsense world framework for trajectory planning in autonomous driving. At the core of C-TRAIL is a human-inspired \textbf{Commonsense World} which interacts with the environment through a closed-loop of \textbf{Recall}, \textbf{Plan}, and \textbf{Update}. It is implemented via three coupled modules, each tackling a core challenge. First, a \rev{Trust-Aware} Commonsense Recall module leverages LLMs to extract structured commonsense from sensor inputs, and employs a trust mechanism to filter and score the recalled information. It ensures only reliable commonsense knowledge is used for planning. Second, the \rev{Trust-Guided Planning} module integrates these trust representations into a Dirichlet trust policy that guides MCTS. This mechanism enables the planner to regulate the degree of reliance on LLM outputs. Finally, a \rev{Trust Calibration Update} module \rev{continuously refines the trust scores and policy parameters based on environmental feedback}, enabling continuous and robust trajectory planning. Through repeated cycles of Recall, Plan, and Update, the Commonsense World continuously interacts with the environment to enable adaptive trajectory planning in a dynamic environment.

Our main contributions are summarized as follows:

\begin{itemize}
    \item To the best of our knowledge, C-TRAIL is the first framework to introduce a human-inspired \textit{Commonsense World} that engages in trajectory planning for autonomous driving via a closed-loop \rev{Recall, Plan, and Update cycle}.

    \item \rev{We propose C-TRAIL with three coupled modules: Trust-Aware Commonsense Recall, Trust-Guided Planning, and Trust Calibration Update, which jointly address reliable commonsense extraction, trust-weighted planning, and adaptive trust refinement.}

    \item \rev{Experiments on four Highway-env scenarios and two real-world levelXData datasets (highD, rounD) show that C-TRAIL reduces ADE by 40.2\%, FDE by 51.7\%, and improves SR by 16.9\,pp over state-of-the-art baselines on average.}
\end{itemize}
\section{Preliminaries}
\label{sec_preli}

\begin{revised}
\subsection{Definition}
\label{sec_commonsense_world}

\noindent\textbf{Definition 1 (State).}
$\mathcal{S}_t = \langle \mathcal{V}_t, \mathcal{E}_t \rangle$ encodes the traffic scene as a graph centered on the ego vehicle. The node set $\mathcal{V}_t$ represents the ego vehicle and $N$ surrounding vehicles:
\begin{equation}
\label{for_kinematic}
    \mathcal{V}_t = \{\, v_{0,t}\,\} \cup \{\, v_{i,t} \mid i \in \{1,\ldots,N\}\,\}
\end{equation}
where each $v_{i,t}$ is a kinematic feature vector comprising position, speed, acceleration, heading, and lane ID. The edge set captures LLM-inferred semantic relations:
\begin{equation}
\label{eq:edge_set}
    \mathcal{E}_t = \{(v_{0,t},\, v_{i,t},\, e_{0,i,t})\}_{i=1}^{N}
\end{equation}
where each $e_{0,i,t} \in \mathcal{R}$ belongs to one of eight spatial types: $\mathcal{R} = \{$\emph{Ahead}, \emph{Back}, \emph{LeftAhead}, \emph{LeftBack}, \emph{Right}, \emph{Left}, \emph{RightAhead}, \emph{RightBack}$\}$.

\noindent\textbf{Definition 2 (Action).}
The ego vehicle selects from a discrete meta-action set $\mathcal{A}=\{$IDLE, Faster, Slower, Turn\_left, Turn\_right$\}$, representing high-level driving decisions applicable across both simulated and real-world scenarios.

\noindent\textbf{Definition 3 (Trust).}
$\mathcal{C}_t = \{c_{0,i,t}\}_{i=1}^{N}$ quantifies the reliability of each relation. Trust is assessed along two dimensions: (1) \emph{commonsense trust} $c_{t,\text{\textit{llm}}}$, evaluating semantic consistency across multiple LLM queries, and (2) \emph{kinematic trust} $c_{t,\text{\textit{kin}}}$, verifying physical feasibility. A scalar aggregate $C_t$ is derived for policy modulation (Section~\ref{Method_trustcommonsenserecall}, Eq.~\ref{eq:combined_trust}).

\noindent\textbf{Definition 4 (Commonsense Graph).}
The commonsense graph $\mathcal{G}_t = (\mathcal{V}_t, \mathcal{E}_t, \mathcal{C}_t)$ is defined by the vehicle nodes $\mathcal{V}_t$, semantic edges $\mathcal{E}_t$, and per-relation trust scores $\mathcal{C}_t$, forming a trust-weighted representation that integrates LLM-derived commonsense with reliability assessment.

\noindent\textbf{Definition 5 (Commonsense World).}
The commonsense world $\mathcal{W}_t$ is defined by the commonsense graph $\mathcal{G}_t$, the action set $\mathcal{A}$, and the transition function $\mathcal{T}$:
\begin{equation}
\label{eq:commonsense_world}
    \mathcal{W}_t = (\mathcal{G}_t, \mathcal{A}, \mathcal{T})
\end{equation}
It provides a unified representation that couples trust-weighted scene understanding with decision-making. The transition $\mathcal{T}$ models the environment dynamics: given $\mathcal{V}_t$ and action $a_t \in \mathcal{A}$, the environment evolves as $(\mathcal{V}_{t+1}, r_t) \sim \mathcal{T}(\mathcal{V}_t, a_t)$ with scalar reward $r_t$. At each new step, $\mathcal{G}_{t+1}$ is reconstructed from updated observations via the Recall--Plan--Update loop.

\subsection{Problem Definition}

We formulate trajectory planning as a \emph{commonsense-guided policy learning} problem. At each step $t$, the agent operates within $\mathcal{W}_t$, which provides trust-weighted LLM commonsense for action selection. The objective is to learn a policy $\pi$ maximizing the cumulative expected reward:
\begin{equation}
\label{eq:objective}
\begin{split}
    \pi^* = \arg\max_{\pi}\;\mathbb{E}_{\pi}\!\left[\sum_{i=0}^{\infty} \gamma^i\, R(\mathcal{W}_{t+i},\, a_{t+i}) \;\middle|\; \mathcal{W}_t\right],\\
    a_{t+i} \sim \pi(\mathcal{W}_{t+i})
\end{split}
\end{equation}
where $R(\mathcal{W}_t, a_t)$ is the reward received after taking action $a_t$ in world $\mathcal{W}_t$, and $\gamma \in [0,1)$ is the discount factor balancing immediate and future rewards.
\end{revised}

\section{Model}
\label{sec_method}
\begin{figure*}
    \centering
    \includegraphics[width=\linewidth]{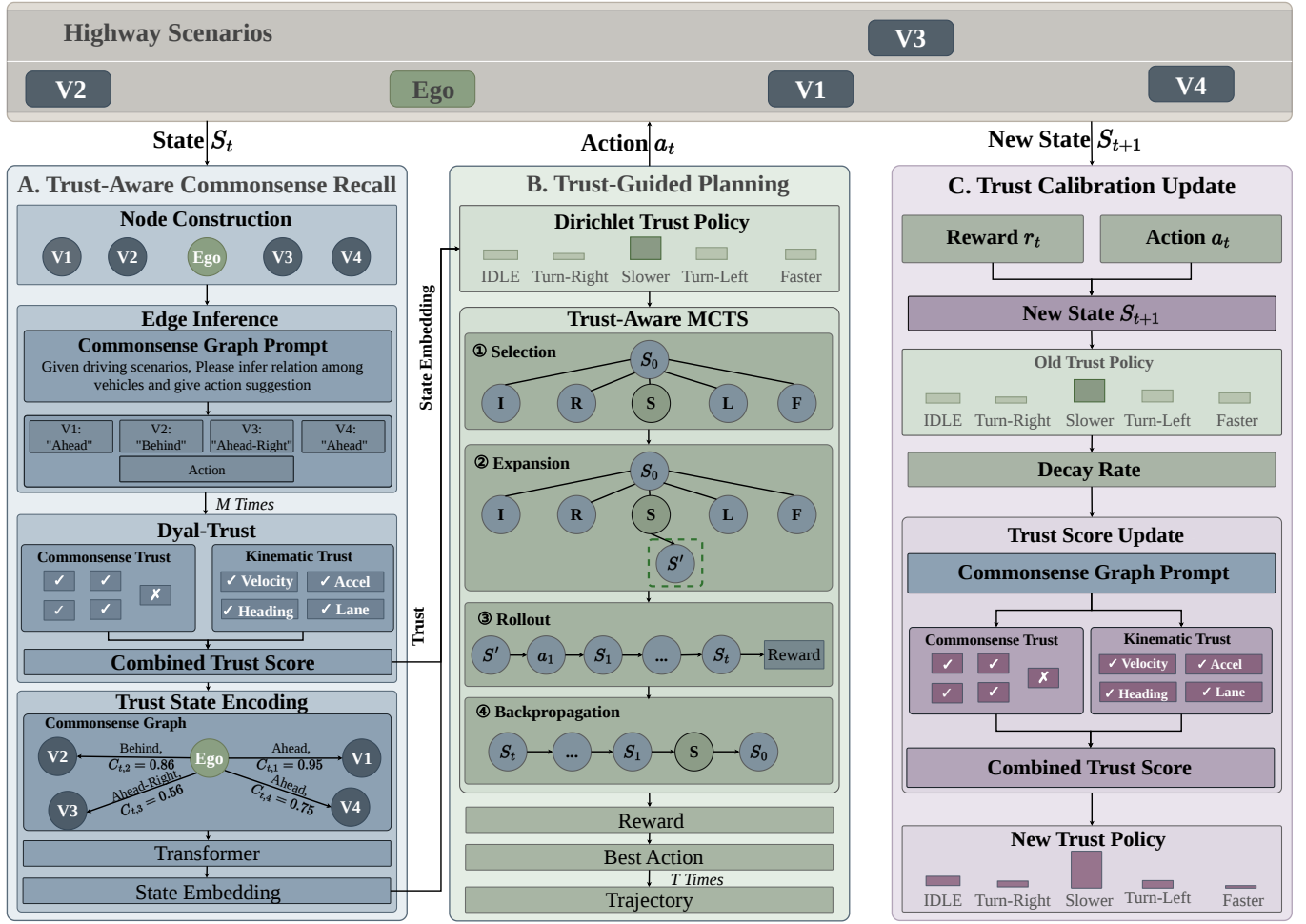}
    \caption{\rev{The C-TRAIL, A Commonsense World Framework for Trajectory Planning in Autonomous Driving}}
    \label{framework}
\end{figure*}

\rev{As shown in Fig.~\ref{framework}, the framework evolves trajectory planning through a closed-loop \textbf{Recall--Plan--Update} cycle: the \emph{Recall} module queries an LLM to extract commonsense and quantifies its reliability; the \emph{Plan} module injects the trust-weighted commonsense into MCTS to guide action selection; and the \emph{Trust Calibration Update} module refines trust scores and policy parameters based on environmental feedback. These modules jointly form a closed loop yielding adaptive trajectories.}\looseness=-1

\begin{figure}[htp]
\centering
\fcolorbox{black}{gray!8}{%
\parbox{0.96\linewidth}{%
\textbf{Commonsense Graph Prompt}

\vspace{4pt}
\footnotesize
[\textbf{TASK}]\\
You are given a structured description of a driving environment, including the driving scenario, the ego vehicle's intention, and a predefined set of high-level actions and relations. Your task is to analyze the provided context and generate two types of outputs:\\
$\bullet$ \textbf{[Action]} A recommended high-level driving action selected from the available action set.\\
$\bullet$ \textbf{[Relations]} A structured set of spatial relations between the ego vehicle and neighboring vehicles, using only the predefined relation types.

[\textbf{DRIVING SCENARIO}]\\
The driving scenario is a four-lane highway.\\
$\bullet$ Ego Vehicle [659]: The ego vehicle is driving in lane 1. Its position is (408.13, 0.00), with a velocity of 24.03 m/s and an acceleration of $-$0.05 m/s\textsuperscript{2}.\\
$\bullet$ Vehicle [992]: This vehicle is positioned ahead and right of the ego vehicle. Its position is (422.72, 4.00), with a velocity of 17.25 m/s and an acceleration of $-$1.33 m/s\textsuperscript{2}.\\
$\bullet$ Vehicle [712]: This vehicle is driving in the same lane as the ego vehicle and is positioned ahead. Its position is (442.23, 0.00), with a velocity of 22.84 m/s and an acceleration of $-$1.49 m/s\textsuperscript{2}.

[\textbf{INTENTION}]\\
Drive safely and avoid potential collisions.

[\textbf{AVAILABLE ACTIONS}]\\
Exactly one high-level action must be selected from:\\
\texttt{IDLE | Turn-Right | Turn-Left | Faster | Slower}

[\textbf{AVAILABLE RELATIONS}]\\
Each relation should be selected from:\\
\texttt{Ahead | Back | Left | Right | LeftAhead | RightAhead | LeftBack | RightBack}

[\textbf{RESPONSE FORMAT}]\\
\texttt{Action: ActionName}\\
\texttt{Relation: [(Ego Vehicle ID, Surrounding Vehicle ID, Relation)]}
}%
}
\caption{Commonsense Graph Prompt.}
\label{fig:commonsense_prompt}
\end{figure}
\subsection{\rev{Trust-Aware} Commonsense Recall}
\label{Method_trustcommonsenserecall}

\rev{This module constructs the commonsense graph $\mathcal{G}_t = (\mathcal{V}_t, \mathcal{E}_t, \mathcal{C}_t)$ (Section~\ref{sec_preli}) from the driving state $\mathcal{S}_t$ by querying an LLM for semantic relations and quantifying their reliability through a dual-trust mechanism.}

\noindent\textbf{\rev{Node Construction.}}
\rev{The node set $\mathcal{V}_t$ represents the ego vehicle and $N$ surrounding vehicles, where each node $v_{i,t}$ encodes sensor-derived kinematic features such as position, speed, acceleration, heading, and lane ID (Eq.~\ref{for_kinematic}).}

\noindent\textbf{\rev{Edge Inference.}}
\rev{The edge set $\mathcal{E}_t$ captures spatial relations between nodes (e.g., \emph{Ahead}, \emph{LeftBack}), which are inferred by an LLM. As illustrated in Fig.~\ref{fig:commonsense_prompt}, the sensor-derived vehicle states are organized into a structured prompt that specifies the driving task, describes the current scenario and vehicle kinematics, and enumerates the available actions and spatial relations. The LLM processes this prompt and returns predicted relations $e = \{e_{0,i,t}\}_{i=1}^{N}$ along with a recommended action $a \in \mathcal{A}$, i.e., $(e, a) \leftarrow \text{LLM}(\text{prompt})$. To improve reliability, we perform $M$ independent queries with temperature sampling and aggregate the results; the LLM is accessed via API without fine-tuning, preserving its general commonsense capabilities. However, LLM outputs are not always reliable. Fig.~\ref{fig:Correct_Output_Example} shows a valid output with consistent relations and a feasible action, while Fig.~\ref{fig:Incorrect_Output_Example} reveals common failure modes such as inconsistent relations and physically infeasible actions.}

\noindent\textbf{\rev{Dual-Trust} Quantification.}
To quantify this unreliability, we introduce a dual-trust mechanism that yields per-relation trust scores $\mathcal{C}_t$, the third component of $\mathcal{G}_t$. It assesses LLM outputs along two complementary dimensions: \emph{commonsense trust} $c_{t,\text{\textit{llm}}}$, which measures semantic consistency across multiple LLM queries, and \emph{kinematic trust} $c_{t,\text{\textit{kin}}}$, which verifies physical feasibility against observed vehicle dynamics.

\noindent\textit{Commonsense Trust.} The commonsense trust $c_{t,\text{\textit{llm}}}$ evaluates the consistency of LLM predictions across $M$ independent queries. For each surrounding vehicle $i$, we obtain $M$ relation predictions $\{e^1_{0,i,t}, ..., e^M_{0,i,t}\}$ and compute:
\begin{equation}
\label{eq:llm_trust_entropy}
\rev{\begin{aligned}
    p_k &= \tfrac{1}{M}\textstyle\sum_{m=1}^{M}\mathbb{I}(e^m_{0,i,t} = k) \\
    H(p) &= -\textstyle\sum_{k} p_k \log p_k \\
    c_{0,i,t,\text{\textit{llm}}} &= \max_{k \in \mathcal{E}} p_k - \alpha_{\textit{llm}} \cdot H(p) \\
    c_{t,\text{\textit{llm}}} &= \tfrac{1}{N}\textstyle\sum_{i=1}^{N} c_{0,i,t,\text{\textit{llm}}}
\end{aligned}}
\end{equation}
where $p_k$ is the empirical probability of relation type $k \in \mathcal{E}$, $\max_{k} p_k$ reflects prediction confidence, $H(p)$ penalizes prediction inconsistency, and $\alpha_{\textit{llm}}$ scales the entropy penalty.

\noindent\textit{Kinematic Trust.} The kinematic trust $c_{t,\text{\textit{kin}}}$ verifies that the recommended action $a$ is physically feasible given the current vehicle state. It incorporates four constraints:
\begin{equation}
\label{eq:kin_trust}
\begin{aligned}
c_{t,\text{\textit{kin}}} = \;&\omega_1\!\left(1-\frac{|\Delta_v|}{v_{\text{max}}}\right) + \omega_2 e^{-\lambda_{\text{acc}}|\Delta_{\text{acc}}|} \\
&+ \omega_3 \cos(\Delta_{\theta}) + \omega_4 L
\end{aligned}
\end{equation}
where $\Delta_v$, $\Delta_{\text{acc}}$, and $\Delta_\theta$ denote the velocity, acceleration, and heading angle differences between the predicted next state (if action $a$ is executed) and the actual observed state, and $L \in \{0, 1\}$ indicates lane consistency. The weights $\omega_1, ..., \omega_4$ are equally weighted.

\noindent\textit{\rev{Combined Trust Score.}} \rev{A semantically consistent prediction is still unreliable if the recommended action violates physical constraints. We therefore integrate both components through a gating mechanism that uses kinematic trust as a prerequisite:}
\begin{equation}
\label{eq:combined_trust}
    C_t = c_{t,\text{\textit{llm}}} \cdot \sigma\big(\kappa(c_{t,\text{\textit{kin}}} - \eta)\big)
\end{equation}
\rev{where $\sigma(\cdot)$ is the sigmoid function, $\eta$ is the feasibility threshold, and $\kappa$ controls the transition steepness. This design ensures that commonsense knowledge is only trusted when it is physically feasible: when $c_{t,\text{\textit{kin}}} \gg \eta$, the commonsense trust is fully retained; when $c_{t,\text{\textit{kin}}} < \eta$, it is suppressed, effectively filtering out unreliable LLM outputs.}

\noindent\textbf{Trust State Encoding.}
\rev{With the commonsense graph $\mathcal{G}_t$ established, the remaining step is to produce a compact representation suitable for MCTS. We encode the trust-weighted graph into a dense state vector using a transformer encoder:}
\begin{equation}
\label{eq:trust_encoding}
\rev{\begin{aligned}
    \textbf{Z}_t &= f_{\text{enc}}(\mathcal{G}_t) = \text{Attn}\bigl(\textbf{v}_{0,t}, \{\textbf{v}_{i,t} {+} \tilde{\textbf{e}}_{0,i,t}\}_{i=1}^{N}\bigr) \\
    \textbf{v}_{i,t} &= \text{MLP}(v_{i,t}), \quad \tilde{\textbf{e}}_{0,i,t} = c_{0,i,t} \cdot \textbf{e}_{0,i,t}
\end{aligned}}
\end{equation}
\rev{where $\textbf{v}_{i,t}$ is the embedding of each vehicle's kinematic state, and $\tilde{\textbf{e}}_{0,i,t}$ is the trust-weighted relation embedding. Attention enables the ego to attend to surrounding vehicles via trust-weighted relations.}\looseness=-1

\begin{figure}[htp]
\centering

\subfigure[Correct output]{%
\begin{minipage}{0.96\linewidth}
\fcolorbox{black}{gray!10}{%
\parbox{0.96\linewidth}{%
\textbf{Correct Output Example}

\vspace{3pt}
\scriptsize
\texttt{Action: Turn-Left}\\[2pt]
\texttt{Relation: [(659, 992, RightAhead), (659, 712, Ahead)]}
}%
}
\end{minipage}
\label{fig:Correct_Output_Example}
}\\[0.3em]

\subfigure[Incorrect output]{%
\begin{minipage}{0.96\linewidth}
\fcolorbox{black}{gray!10}{%
\parbox{0.96\linewidth}{%
\textbf{Incorrect Output Examples}

\vspace{3pt}
\scriptsize

[\textbf{E1}] \textbf{Format Error -- Missing field}\\
\texttt{Turn-Left}\\
\texttt{[(992, RightAhead), (712, Ahead)]}

\smallskip\hrule\smallskip

[\textbf{E2}] \textbf{Invalid Action -- Not in allowed action list}\\
\texttt{Action: Move-Straight}\\
\texttt{Relation: [(659, 992, RightAhead)]}

\smallskip\hrule\smallskip

[\textbf{E3}] \textbf{Tuple Order Error -- Ego and other vehicle reversed}\\
\texttt{Action: Turn-Right}\\
\texttt{Relation: [(992, 659, RightAhead)]}

\smallskip\hrule\smallskip

[\textbf{E4}] \textbf{Illegal Relation Type -- Not in vocabulary}\\
\texttt{Action: Faster}\\
\texttt{Relation: [(659, 992, TopLeft)]}

\smallskip\hrule\smallskip

[\textbf{E5}] \textbf{Multiple Actions Specified -- Only one allowed}\\
\texttt{Action: Turn-Left, Faster}\\
\texttt{Relation: [(659, 992, RightAhead)]}
}%
}
\end{minipage}
\label{fig:Incorrect_Output_Example}
}

\caption{Examples of correct and incorrect LLM outputs.}
\label{fig:Output_Examples}
\end{figure}

\subsection{\rev{Trust-Guided Planning}}
\label{sec:planning}

\rev{Given the graph encoding $\textbf{Z}_t$ and trust score $C_t$ from the Recall module, this module selects an action $a_t$ via MCTS by constructing a Dirichlet trust policy from LLM recommendations and injecting it into a modified PUCT selection rule, where $C_t$ controls the balance between commonsense guidance and exploration.}

\noindent\textbf{Dirichlet Trust Policy.}
\label{sec:dirichlet_policy}
\rev{At each time step $t$, the LLM queries and results from the Recall module are cached and reused across all $K$ MCTS simulations. To convert these discrete LLM outputs into a continuous action prior suitable for tree search, we adopt a Dirichlet distribution inspired by AlphaZero~\cite{silver2018general}, whose concentration parameters depend on the LLM action frequencies and the trust score $C_t$. This formulation provides a natural trust-modulated mapping: high $C_t$ yields a peaked distribution that exploits LLM recommendations, while low $C_t$ yields a flat distribution that encourages exploration. Formally, the Dirichlet trust policy is defined as:}
\begin{equation}
\label{eq:dirichlet_policy}
\rev{\begin{aligned}
    \pi(a \mid \textbf{Z}_t, C_t) &\sim \text{Dirichlet}(\boldsymbol{\alpha}_t), \quad \forall a \in \mathcal{A} \\
    \tilde{\alpha}_t(a) &= f(a) \cdot (1 + \beta \cdot C_t) \\
    f(a) &= \tfrac{1}{M}\textstyle\sum_{m=1}^{M}\mathbb{I}(a^m{=}a) + \epsilon
\end{aligned}}
\end{equation}
\rev{where $f(a)$ is the action prior derived from $M$ LLM queries, with $\epsilon$ providing additive smoothing for unsampled actions. The hyperparameter $\beta$ controls how trust modulates the policy concentration: when $C_t$ is high, the concentration parameters become large, making the policy \emph{peaked} around high-$f(a)$ actions (exploitation); when $C_t$ is low, they become small, \emph{flattening} the policy to encourage exploration. At $t{=}0$, we set $\boldsymbol{\alpha}_0 = \tilde{\boldsymbol{\alpha}}_0$; at subsequent steps, $\boldsymbol{\alpha}_t$ is obtained via exponential moving average from the Update module (Eq.~\ref{eq:policy_update}).}

\noindent\textbf{\rev{Trust-Aware MCTS.}}
\label{sec:trust_mcts}
\rev{The Dirichlet trust policy is injected into MCTS through a modified selection rule, so that unreliable LLM guidance is automatically down-weighted during search.} 

\rev{The procedure (Algorithm~\ref{alg:c-mcts}) runs $K$ simulations per planning cycle, consisting of four phases. During \textbf{\textit{Selection}}, the tree is traversed from the root by choosing at each node $h$ the action that maximizes a trust-aware PUCT selection:}
\begin{equation}
\label{eq:puct}
\rev{\begin{split}
a^* = \arg\max_{a \in \mathcal{A}} \bigl[ Q(h, a) & \\
  + \;\lambda \cdot C_t \cdot \pi(a|\textbf{Z}_t, C_t) &\cdot \frac{\sqrt{N(h)}}{1 + N(h,a)} \bigr]
\end{split}}
\end{equation}
\rev{where $Q(h, a)$ is the estimated action value, $\pi(a|\textbf{Z}_t, C_t)$ is the Dirichlet trust policy, $\frac{\sqrt{N(h)}}{1 + N(h,a)}$ is the UCB exploration bonus, and $\lambda$ is the exploration constant. The key design is that $C_t$ multiplies the prior term: when trust is high, the commonsense prior strongly guides the search; when trust is low, the search relies more on Q-values and exploration. Upon reaching an unexpanded node, \textbf{\textit{Expansion}} adds it to the tree and initializes its statistics ($N$, $Q$). \textbf{\textit{Rollout}} then simulates from the new node to a terminal state or depth cutoff using a uniform random policy. The resulting return is \textbf{\textit{Backpropagated}} along the traversed path to update $Q(h,a)$ and $N(h,a)$. After $K$ simulations, the final action $a_t$ is selected by visit count $N(h_0, a)$. The MCTS tree is reinitialized at each time step with updated Dirichlet parameters from the Update module (Section~\ref{sec:policy_update}); updates thus occur online at every planning cycle rather than only between episodes.}

\begin{algorithm}
\begin{revised}
\caption{Trust-Guided Monte Carlo Planning}
\label{alg:c-mcts}
\begin{algorithmic}[1]
\STATE \textbf{Input:} $\textbf{Z}_t,\, C_t,\, f(a),\, K,\, \tau$
\STATE \textbf{Output:} Optimal action $a_t$
\vspace{2pt}
\STATE \textbf{PLAN}($\textbf{Z}_t, C_t, f(a), K$):
\STATE $\boldsymbol{\alpha}_t \!\leftarrow\! \{f(a)(1 {+} \beta C_t)\}_{a \in \mathcal{A}}$
\STATE $\pi(\cdot|\textbf{Z}_t, C_t) \sim \text{Dirichlet}(\boldsymbol{\alpha}_t)$
\STATE Initialize tree $\Gamma$ with root $h_0$
\FOR{$k = 1$ to $K$}
    \STATE SIMULATE($s_t, h_0, 0, \Gamma$)
\ENDFOR
\RETURN $a_t = \arg\max_{a} N(h_0, a)$
\vspace{2pt}
\STATE \textbf{SIMULATE}($s, h, d, \Gamma$):
\IF{$\gamma^d < \tau$ \textbf{or} terminal($s$)}
    \RETURN 0
\ENDIF
\IF{$h \notin \Gamma$}
    \STATE \textit{Expansion:} $\Gamma \!\leftarrow\! \Gamma \cup \{h\}$;\; $N(h)\!\leftarrow\!0$,\; $Q(h,a)\!\leftarrow\!0\;\forall a$
    \STATE \textit{Rollout:} $a_k \!\sim\! \text{Uniform}(\mathcal{A})$,\; $(s_{k+1}, r_k) \!\sim\! \mathcal{T}(s_k, a_k)$
    \RETURN $\textstyle\sum_{k=0}^{T} \gamma^k r_k$
\ENDIF
\STATE \textit{Selection:} $a^* \!\leftarrow\! \arg\max_{a}\bigl[Q(h,a) \!+\! \lambda C_t \pi(a)\tfrac{\sqrt{N(h)}}{1\!+\!N(h,a)}\bigr]$
\STATE $(s', r) \sim \mathcal{T}(s, a^*)$;\; $R \leftarrow r + \gamma \cdot \text{SIMULATE}(s', h {\cup} \{a^*\}, d{+}1, \Gamma)$
\STATE \textit{Backpropagation:} $N(h, a^*) \leftarrow N(h, a^*) + 1$
\STATE $Q(h, a^*) \leftarrow Q(h, a^*) + \frac{R - Q(h, a^*)}{N(h, a^*)}$
\RETURN $R$
\end{algorithmic}
\end{revised}
\end{algorithm}

\noindent\textbf{\rev{Trajectory Generation.}}
\rev{By repeating the Recall--Plan--Update loop at each decision step, C-TRAIL generates a trajectory $\tau = \{a_t, a_{t+1}, \ldots, a_{t+T}\}$ in a receding-horizon fashion, where each action leverages updated trust scores and scene context.}\looseness=-1

\subsection{\rev{Trust Calibration Update}}
\label{sec:adaptive_update}

\rev{Once the selected action $a_t$ is executed, the environment provides two feedback signals: (i)~new vehicle states $\mathcal{S}_{t+1} = \{s_{i,t+1}\}$, which reveal actual positions, velocities, and headings of surrounding vehicles and thereby enable reassessment of LLM-predicted relations; and (ii)~a scalar task reward $r_t$, which reflects the overall quality of the executed action. This module uses both signals to recalibrate trust and policy online, closing the Recall--Plan--Update loop: the new states drive trust recomputation via masked blending, while the reward modulates the rate at which stale trust decays.}

\noindent\textbf{Trust Score Update.}
\label{sec:trust_calibration}
\rev{Using the new vehicle states $\mathcal{S}_{t+1}$, we recompute commonsense trust (Eq.~\ref{eq:llm_trust_entropy}) and kinematic trust (Eq.~\ref{eq:kin_trust}) for each observed vehicle, yielding fresh estimates $\tilde{c}^{t+1}$. These are blended with historical values via an observation mask $m_{i,t+1} \in \{0,1\}$ (1 if vehicle $i$ is detected, 0 otherwise). For commonsense trust, we additionally apply a reward-adaptive decay toward a neutral baseline $c_0$ to prevent overconfidence in stale LLM outputs:}
\begin{equation}
\label{eq:trust_update}
\rev{\begin{aligned}
    c^{t+1}_{0,i,\text{llm}} &= \gamma_t \bigl[
      m_{i,t+1} \tilde{c}^{t+1}_{0,i,\text{llm}} +
      (1{-}m_{i,t+1}) c^{t}_{0,i,\text{llm}}
    \bigr] \\
    &\quad + (1{-}\gamma_t) c_0 \\
    c^{t+1}_{0,i,\text{kin}} &=
      m_{i,t+1} \tilde{c}^{t+1}_{0,i,\text{kin}} +
      (1{-}m_{i,t+1}) c^{t}_{0,i,\text{kin}} \\
    \gamma_t &= \gamma_{\text{decay}} +
      (1{-}\gamma_{\text{decay}}) \sigma(\kappa_r r_t)
\end{aligned}}
\end{equation}
\rev{where $\tilde{c}^{t+1}$ denotes trust recomputed from $\mathcal{S}_{t+1}$ via Eqs.~(\ref{eq:llm_trust_entropy})--(\ref{eq:kin_trust}), $\gamma_{\text{decay}}$ is the base decay rate, $c_0$ is a neutral baseline, and $\kappa_r$ controls reward sensitivity. When $r_t$ is high, $\gamma_t \to 1$ and commonsense trust is largely preserved; when $r_t$ is low (collision, discomfort), $\gamma_t$ drops toward $\gamma_{\text{decay}}$, accelerating decay toward the baseline. Kinematic trust is not decayed because it is always recomputed from physical measurements. The combined trust $C_{t+1}$ is then obtained via Eq.~(\ref{eq:combined_trust}).}

\noindent\textbf{Dirichlet Policy Update.}
\label{sec:policy_update}
\rev{With the recalibrated trust score $C_{t+1}$, we update the Dirichlet concentration parameters via exponential moving average to maintain temporal consistency:}
\begin{equation}
\label{eq:policy_update}
\rev{\begin{aligned}
    \tilde{\alpha}_{t+1}(a) &= f(a) \cdot (1 + \beta \cdot C_{t+1}) \\
    \alpha_{t+1}(a) &= (1 - \gamma_{\text{diri}}) \cdot \alpha_t(a) + \gamma_{\text{diri}} \cdot \tilde{\alpha}_{t+1}(a)
\end{aligned}}
\end{equation}
\rev{where $\tilde{\alpha}_{t+1}(a)$ is the fresh target from updated trust (Eq.~\ref{eq:dirichlet_policy}), and $\gamma_{\text{diri}} \in [0,1]$ controls the update rate. This EMA mechanism ensures temporal consistency: the policy evolves smoothly rather than resetting abruptly at each step, requiring no gradient computation during deployment. The updated $\mathcal{G}_{t+1}$ and $\boldsymbol{\alpha}_{t+1}$ are carried forward to the next Recall cycle at $t+1$.}

\subsection{\rev{Joint Training}}
\label{sec:training}

\rev{The joint training objective combines policy optimization and trust, formulated as:}
\begin{equation}
\label{eq:total_loss}
    \mathcal{L}_{\text{total}} = \lambda_{\text{trust}} \mathcal{L}_{\text{trust}} + \lambda_{\text{policy}} \mathcal{L}_{\text{policy}}
\end{equation}
where $\lambda_{\text{trust}}$ and $\lambda_{\text{policy}}$ are balancing hyperparameters.

\noindent\textbf{Trust Loss.}
\label{sec:trust_loss}
\rev{The trust loss recalibrates the encoder by penalizing value prediction errors in proportion to the trust score:}
\begin{equation}
\label{eq:trust_loss}
\rev{\begin{aligned}
    R(h,a) &= \textstyle\sum_{k=0}^{T} \gamma^k r_{t+k} \\
    \delta_t(h,a) &= Q(h,a) - R(h,a) \\
    \mathcal{L}_{\text{trust}} &= \mathbb{E}_{(h,a) \sim \pi} \left[ C_t \cdot \delta_t(h,a)^2 \right]
\end{aligned}}
\end{equation}
\rev{where $R(h,a)$ is the observed cumulative return, $\delta_t$ is the value prediction error, and $C_t$ weights the squared error. When the encoder assigns high $C_t$ but $|\delta_t|$ is large, a strong gradient pushes the encoder to lower its trust output; when $C_t$ is already low, the same error incurs a small penalty because the system was already skeptical.}

\noindent\textbf{Policy Loss.}
\label{sec:policy_loss}
\rev{The policy loss uses a trust-weighted proximal policy objective to align the Dirichlet trust policy with action values discovered by MCTS:}
\begin{equation}
\label{eq:policy_loss}
\rev{\begin{aligned}
    \mathcal{L}_{\text{policy}} &= \mathbb{E}_{(h,a)} \bigl[
     -C_t \min\!\bigl(\rho_t A(h,a),\;
      \text{clip}(\rho_t, 1{-}\epsilon, \\
      &\quad 1{+}\epsilon) A(h,a)\bigr)
     + \lambda_{\text{KL}} D_{\text{KL}}\bigl(\hat{\pi} \| \pi_Q\bigr)
    \bigr]
\end{aligned}}
\end{equation}
\rev{where $A(h,a)$ is the advantage function, $\rho_t = \hat{\pi}(a|h)/\hat{\pi}_{\text{old}}(a|h)$ is the importance sampling ratio, $\hat{\pi}(a|h)$ is the Dirichlet trust policy, $\pi_Q(a|h)$ is the soft policy induced from MCTS Q-values, and $\epsilon$ is the PPO clipping threshold. $C_t$ scales the clipped surrogate so that policy updates are stronger when commonsense is reliable. The KL term regularizes $\hat{\pi}$ toward $\pi_Q$, keeping the Dirichlet prior close to MCTS discoveries.}

\begin{revised}
\subsection{Deployment Architecture}
\label{sec:deployment}

C-TRAIL is designed as a \emph{high-level strategic planner} that operates in a receding-horizon fashion: it generates a multi-step trajectory plan every 5--10\,s and delegates execution to a fast downstream controller (e.g., PID or MPC) running at the standard 100\,ms control loop. Its input interface consumes structured kinematic states (position, velocity, heading, lane~ID) from upstream perception modules, and the output is a sequence of discrete meta-actions translatable to continuous control commands. The LLM query in the Recall stage is invoked asynchronously once per planning cycle rather than during MCTS simulations, and response caching or locally deployed models (e.g., LLaMA) ensure compatibility with real-time constraints. When LLM latency exceeds a predefined threshold or trust scores fall below $\eta$, the planner gracefully degrades to exploration-dominant MCTS without commonsense guidance, ensuring safety continuity. This modular design allows C-TRAIL to be integrated into standard AV software stacks such as Autoware or Apollo.
\end{revised}

\section{Experiments}
\subsection{Overall Settings}

\noindent\textbf{Data.}
\label{sec:data_collection}
We evaluate C-TRAIL in both simulated and real-world settings. 1)~\textit{\textbf{Simulated environments.}} We adopt the Highway-env simulator~\cite{highway-env}, a widely used platform in LLM-based driving research~\cite{wen2023dilu,cui2024survey,xi2022graph}. It provides configurable traffic scenarios with tunable lane count, vehicle density, and road topology, satisfying the core requirements for trajectory planning evaluation. We select four representative scenarios: \emph{highway}, \emph{merge}, \emph{roundabout}, and \emph{intersection}. Following prior benchmarks~\cite{wen2023dilu,xi2022graph}, each episode runs 10 steps at 1\,Hz; we collect $10^5$ step-level samples per scenario. \rev{2)~\textit{\textbf{Real-world datasets.}} We further evaluate on two drone-captured trajectory datasets from the German levelXData collection: \emph{highD}~\cite{highDdataset} and \emph{rounD}~\cite{rounDdataset}. These datasets provide naturalistic driving trajectories with measurement noise and partial occlusion absent from simulation, enabling us to test whether the framework transfers beyond controlled environments. We extract trajectory segments of 10 steps at 1\,Hz and apply the same kinematic feature extraction and LLM prompting pipeline.}

\begin{revised}
To assess whether models can generalize to novel traffic conditions beyond their training distribution, we define two evaluation protocols.
1)~\textit{\textbf{Seen}} environments adopt the default configurations from prior benchmarks~\cite{wen2023dilu,xi2022graph}: 4 lanes for highway/merge, 2 lanes for roundabout/intersection, and vehicle density 2.0.
2)~\textit{\textbf{Unseen}} environments introduce distribution shifts not present during training: the lane count is expanded to 4--5 for highway/merge, and vehicle density is raised to 3.0. All experiments are repeated over 5 random seeds; we report the mean with 95\% confidence intervals. Unless otherwise specified, C-TRAIL refers to the GPT-4o variant. TABLE~\ref{tab:exp_settings} summarizes all hyperparameters.
\end{revised}

\noindent\textbf{Baselines.}
\rev{We compare with seven baselines from three categories:}
\rev{(i) \emph{\textbf{classical non-LLM planners}} (DQN~\cite{mnih2015human,song2025inverse}, MCTS~\cite{coulom2006efficient}, GRAD~\cite{xi2022graph,liu2025mhgin}), evaluating gains beyond purely data-driven or uninformed search-based planning;
(ii) \emph{\textbf{LLM fine-tuned planners}} (LMTraj~\cite{bae2024can}), testing whether fine-tuning an LLM as a trajectory predictor outperforms using it as a commonsense oracle; and (iii) \emph{\textbf{LLM-guided planners}} (DiLu~\cite{wen2023dilu}, LangMPC~\cite{sha2023languagempc}, GPT-Driver~\cite{mao2023gpt}), which incorporate LLM outputs directly without explicit reliability assessment, contrasting with C-TRAIL's trust-aware integration.
All methods use identical observation and action spaces.}

\noindent\textbf{Evaluation Metrics.}
\rev{We adopt four metrics:
\textbf{ADE} (Average Displacement Error) measures overall trajectory accuracy;
\textbf{FDE} (Final Displacement Error) captures endpoint precision critical for goal reaching;
\textbf{SR} (Success Rate, $\delta{=}2$\,m) reflects planning success from a safety perspective; and
\textbf{RPA} (Relation Prediction Accuracy) quantifies the correctness of LLM-generated commonsense before it enters the planner.}

\begin{table}[!t]
\centering
\begin{revised}
\footnotesize
\setlength{\tabcolsep}{1.5pt}
\begin{tabular}{l|l}
\toprule
\textbf{Category} & \textbf{Setting} \\
\midrule
\multicolumn{2}{c}{\textit{Simulation Environment}} \\
\midrule
Simulator & Highway-env~\cite{highway-env} \\
Scenarios & Highway, Merge, Roundabout, \\
 & Intersection \\
Action space $|\mathcal{A}|$ & \{IDLE, FASTER, SLOWER, \\
 & LANE\_LEFT, LANE\_RIGHT\} \\
\midrule
\multicolumn{2}{c}{\textit{Real-world Datasets}} \\
\midrule
highD~\cite{highDdataset} & German highway (drone) \\
rounD~\cite{rounDdataset} & German roundabouts (drone) \\
\midrule
\multicolumn{2}{c}{\textit{Trust-Aware Commonsense Recall}} \\
\midrule
LLM backbone & GPT-3.5-turbo / GPT-4o \\
LLM temperature & 0.7 \\
Max output tokens & 4096 \\
Independent queries $M$ & 5 \\
Text embedding & text-embedding-ada-002 \\
Embedding dimension & 1536 \\
Entropy penalty $\alpha_{\textit{llm}}$ & 0.3 \\
Kinematic weights $\omega_{1\text{-}4}$ & 0.25  \\
Acceleration decay $\lambda_{\textit{acc}}$ & 0.5 \\
Max speed $v_{\text{max}}$ & 40 m/s \\
Architecture & Transformer \\
Encoder layers & 2 \\
Attention heads & 4 \\
Hidden dimension & 128 \\
Dropout rate & 0.1 \\
\midrule
\multicolumn{2}{c}{\textit{Trust-Guided Planning}} \\
\midrule
Simulations per step $K$ & 50 \\
Lookahead depth $d$ & 10 \\
Discount factor $\gamma$ & 0.99 \\
Rollout cutoff $\tau$ & 0.01 \\
Exploration constant $\lambda$ & 1.0 \\
Trust influence $\beta$ & 1.5 \\
Smoothing constant $\epsilon$ & 0.1 \\
Inference time & 14--18 s/step \\
\midrule
\multicolumn{2}{c}{\textit{Trust Calibration Update}} \\
\midrule
Dirichlet update rate $\gamma_{\text{diri}}$ & 0.3 \\
Trust decay rate $\gamma_{\text{decay}}$ & 0.95 \\
Neutral baseline $c_0$ & 0.5 \\
\midrule
\multicolumn{2}{c}{\textit{Hardware \& Training Configuration}} \\
\midrule
Operating system & Ubuntu 22.04 \\
GPU & NVIDIA TITAN RTX (23GB) \\
Optimizer & Adam ($\beta_1$=0.9, $\beta_2$=0.999) \\
Learning rate & $1 \times 10^{-4}$ \\
Batch size & 64 \\
Training epochs & 100 \\
Samples per scenario & $10^5$ step-level \\
Random seeds & 5 \\
\bottomrule
\end{tabular}
\caption{Experiment settings and hyperparameters for C-TRAIL.}
\label{tab:exp_settings}
\end{revised}
\end{table}

\begin{table*}[!t]
\centering
\begin{revised}
\small
\caption{Performance Comparison in Seen and Unseen Environments.}
\label{tab:results}
\renewcommand{\arraystretch}{1.05}
\fontsize{7.5pt}{9pt}\selectfont
\setlength{\tabcolsep}{2.5pt}
\begin{tabular}{ll ccccccccc c}
\toprule
\textbf{Scenario} & \textbf{Metric} & \textbf{DQN} & \textbf{MCTS} & \textbf{GRAD} & \textbf{LMTraj} & \textbf{DiLu} & \textbf{LangMPC} & \textbf{GPT-Dr} & \textbf{C-TRAIL (3.5)} & \textbf{C-TRAIL (4o)} & \textit{p}-value \\
\midrule
\multicolumn{12}{c}{\textit{Seen Environments}} \\
\midrule
\multirow{3}{*}{Highway}
& ADE $\downarrow$ & 2.87$\pm$0.31 & 2.63$\pm$0.26 & 2.51$\pm$0.25 & 1.32$\pm$0.15 & 1.15$\pm$0.15 & 1.25$\pm$0.18 & 1.20$\pm$0.17 & \textbf{0.77$\pm$0.12} & \textbf{0.68$\pm$0.11} & 3.1e-5 \\
& FDE $\downarrow$ & 5.12$\pm$0.27 & 4.82$\pm$0.25 & 4.61$\pm$0.22 & 3.24$\pm$0.18 & 2.81$\pm$0.17 & 3.05$\pm$0.20 & 2.95$\pm$0.19 & \textbf{0.92$\pm$0.20} & \textbf{0.80$\pm$0.09} & 8.7e-6 \\
& SR(\%) $\uparrow$    & 25.4$\pm$3.3  & 29.5$\pm$2.5  & 31.2$\pm$2.0  & 68.3$\pm$2.4  & 73.2$\pm$1.9  & 70.8$\pm$2.2  & 71.5$\pm$2.1  & \textbf{86.6$\pm$2.5}  & \textbf{88.3$\pm$1.6} & 5.2e-6 \\
\midrule
\multirow{3}{*}{Merge}
& ADE $\downarrow$ & 3.72$\pm$0.25 & 3.48$\pm$0.22 & 3.36$\pm$0.18 & 2.17$\pm$0.32 & 1.83$\pm$0.22 & 2.04$\pm$0.28 & 1.95$\pm$0.26 & \textbf{1.27$\pm$0.58} & \textbf{0.93$\pm$0.35} & 4.8e-3 \\
& FDE $\downarrow$ & 5.93$\pm$0.21 & 5.61$\pm$0.18 & 5.47$\pm$0.16 & 3.89$\pm$0.37 & 3.07$\pm$0.36 & 3.52$\pm$0.39 & 3.35$\pm$0.34 & \textbf{1.62$\pm$0.28} & \textbf{1.42$\pm$0.32} & 1.4e-4 \\
& SR(\%) $\uparrow$    & 23.1$\pm$3.7  & 26.8$\pm$3.1  & 28.9$\pm$2.7  & 62.5$\pm$4.6  & 71.3$\pm$2.5  & 66.7$\pm$4.2  & 68.2$\pm$3.9  & \textbf{85.1$\pm$2.9}  & \textbf{87.1$\pm$1.8} & 7.6e-5 \\
\midrule
\multirow{3}{*}{Roundabout}
& ADE $\downarrow$ & 4.15$\pm$0.19 & 3.95$\pm$0.16 & 3.83$\pm$0.11 & 2.52$\pm$0.16 & 2.07$\pm$0.20 & 2.35$\pm$0.18 & 2.28$\pm$0.18 & \textbf{1.53$\pm$0.27} & \textbf{1.38$\pm$0.25} & 2.3e-3 \\
& FDE $\downarrow$ & 5.68$\pm$0.29 & 5.24$\pm$0.25 & 5.09$\pm$0.22 & 3.47$\pm$0.60 & 2.94$\pm$0.33 & 3.28$\pm$0.54 & 3.15$\pm$0.48 & \textbf{2.16$\pm$0.29} & \textbf{1.75$\pm$0.70} & 1.8e-2 \\
& SR(\%) $\uparrow$    & 24.3$\pm$4.8  & 27.6$\pm$4.5  & 29.7$\pm$4.2  & 60.7$\pm$3.2  & 68.2$\pm$3.6  & 63.5$\pm$3.4  & 65.4$\pm$3.2  & \textbf{85.7$\pm$3.7}  & \textbf{85.2$\pm$2.2} & 4.1e-4 \\
\midrule
\multirow{3}{*}{Intersection}
& ADE $\downarrow$ & 5.08$\pm$0.21 & 4.78$\pm$0.18 & 4.62$\pm$0.15 & 3.74$\pm$0.24 & 2.92$\pm$0.27 & 3.41$\pm$0.25 & 3.25$\pm$0.25 & \textbf{1.88$\pm$0.69} & \textbf{1.62$\pm$0.73} & 2.5e-2 \\
& FDE $\downarrow$ & 7.85$\pm$0.48 & 7.48$\pm$0.45 & 7.23$\pm$0.43 & 5.89$\pm$0.32 & 5.05$\pm$0.23 & 5.52$\pm$0.28 & 5.38$\pm$0.26 & \textbf{3.74$\pm$0.46} & \textbf{3.05$\pm$0.34} & 6.3e-5 \\
& SR(\%) $\uparrow$    & 19.6$\pm$5.1  & 22.5$\pm$4.7  & 24.7$\pm$4.5  & 56.2$\pm$4.2  & 64.0$\pm$3.9  & 59.8$\pm$4.0  & 61.3$\pm$3.8  & \textbf{82.3$\pm$3.3}  & \textbf{84.2$\pm$3.2} & 9.4e-5 \\
\midrule
\multirow{3}{*}{highD}
& ADE $\downarrow$ & 3.25$\pm$0.36 & 2.98$\pm$0.32 & 2.84$\pm$0.29 & 1.58$\pm$0.19 & 1.37$\pm$0.18 & 1.49$\pm$0.22 & 1.44$\pm$0.21 & \textbf{0.95$\pm$0.16} & \textbf{0.82$\pm$0.14} & 5.7e-4 \\
& FDE $\downarrow$ & 5.58$\pm$0.33 & 5.21$\pm$0.30 & 5.03$\pm$0.26 & 3.62$\pm$0.23 & 3.15$\pm$0.21 & 3.41$\pm$0.25 & 3.30$\pm$0.24 & \textbf{1.18$\pm$0.24} & \textbf{1.03$\pm$0.13} & 2.1e-5 \\
& SR(\%) $\uparrow$    & 22.1$\pm$3.9  & 26.3$\pm$3.2  & 28.5$\pm$2.5  & 64.8$\pm$2.8  & 70.5$\pm$2.5  & 67.2$\pm$2.6  & 68.6$\pm$2.5  & \textbf{83.4$\pm$2.7}  & \textbf{85.7$\pm$1.9} & 8.3e-5 \\
\midrule
\multirow{3}{*}{rounD}
& ADE $\downarrow$ & 4.53$\pm$0.25 & 4.30$\pm$0.21 & 4.17$\pm$0.17 & 2.81$\pm$0.20 & 2.34$\pm$0.24 & 2.62$\pm$0.22 & 2.53$\pm$0.23 & \textbf{1.72$\pm$0.31} & \textbf{1.56$\pm$0.28} & 3.6e-3 \\
& FDE $\downarrow$ & 6.12$\pm$0.34 & 5.71$\pm$0.31 & 5.52$\pm$0.27 & 3.85$\pm$0.63 & 3.28$\pm$0.38 & 3.61$\pm$0.59 & 3.48$\pm$0.53 & \textbf{2.41$\pm$0.33} & \textbf{2.05$\pm$0.75} & 2.7e-2 \\
& SR(\%) $\uparrow$    & 21.5$\pm$5.3  & 24.8$\pm$4.9  & 27.0$\pm$4.6  & 57.3$\pm$3.6  & 65.1$\pm$3.9  & 60.8$\pm$3.8  & 62.5$\pm$3.6  & \textbf{82.6$\pm$3.9}  & \textbf{83.0$\pm$2.5} & 1.5e-4 \\
\midrule
\multicolumn{12}{c}{\textit{Unseen Environments}} \\
\midrule
\multirow{3}{*}{Highway}
& ADE $\downarrow$ & 3.42$\pm$0.46 & 3.15$\pm$0.42 & 3.94$\pm$0.40 & 1.51$\pm$0.39 & 1.42$\pm$0.57 & 1.48$\pm$0.51 & 1.46$\pm$0.53 & \textbf{0.86$\pm$0.23} & \textbf{0.76$\pm$0.25} & 6.2e-3 \\
& FDE $\downarrow$ & 5.78$\pm$0.39 & 5.38$\pm$0.35 & 5.02$\pm$0.32 & 3.75$\pm$0.25 & 3.62$\pm$0.34 & 3.70$\pm$0.29 & 3.68$\pm$0.31 & \textbf{1.12$\pm$0.34} & \textbf{0.97$\pm$0.24} & 4.5e-5 \\
& SR(\%) $\uparrow$    & 18.7$\pm$4.6  & 22.3$\pm$5.1  & 24.8$\pm$5.7  & 59.2$\pm$5.3  & 64.1$\pm$3.3  & 61.5$\pm$3.7  & 62.3$\pm$3.5  & \textbf{83.8$\pm$3.8}  & \textbf{86.8$\pm$6.0} & 1.3e-4 \\
\midrule
\multirow{3}{*}{Merge}
& ADE $\downarrow$ & 4.18$\pm$0.42 & 3.82$\pm$0.39 & 3.59$\pm$0.46 & 2.62$\pm$0.34 & 2.53$\pm$0.39 & 2.58$\pm$0.37 & 2.55$\pm$0.35 & \textbf{1.36$\pm$0.26} & \textbf{1.07$\pm$0.20} & 7.8e-5 \\
& FDE $\downarrow$ & 6.52$\pm$0.51 & 6.18$\pm$0.54 & 5.94$\pm$0.61 & 4.83$\pm$0.51 & 4.59$\pm$0.35 & 4.72$\pm$0.46 & 4.65$\pm$0.42 & \textbf{1.72$\pm$0.39} & \textbf{1.67$\pm$0.32} & 2.9e-5 \\
& SR(\%) $\uparrow$    & 16.4$\pm$4.2  & 20.1$\pm$3.7  & 22.7$\pm$3.2  & 57.4$\pm$4.3  & 62.5$\pm$2.0  & 59.8$\pm$3.2  & 60.8$\pm$3.1  & \textbf{82.7$\pm$3.0}  & \textbf{85.5$\pm$4.0} & 5.1e-5 \\
\midrule
\multirow{3}{*}{Roundabout}
& ADE $\downarrow$ & 4.85$\pm$0.60 & 4.58$\pm$0.62 & 4.42$\pm$0.66 & 3.24$\pm$0.53 & 2.84$\pm$0.44 & 3.08$\pm$0.50 & 3.02$\pm$0.48 & \textbf{1.62$\pm$0.20} & \textbf{1.52$\pm$0.29} & 1.7e-3 \\
& FDE $\downarrow$ & 6.43$\pm$0.59 & 6.12$\pm$0.60 & 5.94$\pm$0.64 & 4.25$\pm$0.32 & 4.13$\pm$0.32 & 4.20$\pm$0.35 & 4.18$\pm$0.34 & \textbf{2.27$\pm$0.40} & \textbf{2.08$\pm$0.48} & 3.4e-4 \\
& SR(\%) $\uparrow$    & 15.8$\pm$6.4  & 18.2$\pm$6.9  & 20.4$\pm$7.5  & 54.1$\pm$5.0  & 59.8$\pm$1.8  & 56.5$\pm$4.3  & 57.5$\pm$3.9  & \textbf{81.5$\pm$3.2}  & \textbf{84.3$\pm$3.8} & 6.8e-5 \\
\midrule
\multirow{3}{*}{Intersection}
& ADE $\downarrow$ & 5.62$\pm$0.47 & 5.28$\pm$0.46 & 5.05$\pm$0.51 & 4.24$\pm$0.43 & 3.98$\pm$0.42 & 4.12$\pm$0.40 & 4.05$\pm$0.39 & \textbf{2.02$\pm$0.23} & \textbf{1.78$\pm$0.04} & 8.1e-5 \\
& FDE $\downarrow$ & 8.75$\pm$0.53 & 8.42$\pm$0.50 & 8.13$\pm$0.46 & 6.85$\pm$0.31 & 6.92$\pm$0.30 & 6.88$\pm$0.33 & 6.82$\pm$0.32 & \textbf{3.87$\pm$0.33} & \textbf{3.58$\pm$0.06} & 1.2e-6 \\
& SR(\%) $\uparrow$    & 13.5$\pm$6.2  & 16.8$\pm$6.6  & 19.3$\pm$7.6  & 49.7$\pm$3.9  & 57.9$\pm$2.0  & 53.2$\pm$3.3  & 54.8$\pm$3.2  & \textbf{79.4$\pm$2.2}  & \textbf{82.1$\pm$3.2} & 4.7e-5 \\
\midrule
\multirow{3}{*}{highD}
& ADE $\downarrow$ & 3.81$\pm$0.51 & 3.52$\pm$0.47 & 3.35$\pm$0.44 & 1.82$\pm$0.42 & 1.68$\pm$0.61 & 1.76$\pm$0.55 & 1.73$\pm$0.58 & \textbf{1.08$\pm$0.26} & \textbf{0.94$\pm$0.29} & 3.9e-2 \\
& FDE $\downarrow$ & 6.24$\pm$0.46 & 5.82$\pm$0.41 & 5.51$\pm$0.38 & 4.15$\pm$0.30 & 3.96$\pm$0.39 & 4.08$\pm$0.34 & 4.03$\pm$0.36 & \textbf{1.38$\pm$0.37} & \textbf{1.21$\pm$0.27} & 7.2e-5 \\
& SR(\%) $\uparrow$    & 15.3$\pm$5.1  & 19.1$\pm$5.4  & 21.5$\pm$6.1  & 55.6$\pm$5.7  & 61.2$\pm$3.8  & 58.0$\pm$4.2  & 59.2$\pm$3.9  & \textbf{80.5$\pm$4.1}  & \textbf{83.2$\pm$6.3} & 2.4e-4 \\
\midrule
\multirow{3}{*}{rounD}
& ADE $\downarrow$ & 5.21$\pm$0.65 & 4.92$\pm$0.67 & 4.75$\pm$0.70 & 3.52$\pm$0.58 & 3.10$\pm$0.48 & 3.35$\pm$0.54 & 3.28$\pm$0.53 & \textbf{1.85$\pm$0.25} & \textbf{1.71$\pm$0.33} & 5.3e-3 \\
& FDE $\downarrow$ & 6.95$\pm$0.64 & 6.60$\pm$0.66 & 6.38$\pm$0.69 & 4.68$\pm$0.37 & 4.52$\pm$0.38 & 4.61$\pm$0.40 & 4.57$\pm$0.39 & \textbf{2.56$\pm$0.45} & \textbf{2.35$\pm$0.53} & 9.6e-4 \\
& SR(\%) $\uparrow$    & 12.7$\pm$6.8  & 15.4$\pm$7.4  & 17.8$\pm$7.9  & 50.8$\pm$5.4  & 56.5$\pm$2.5  & 53.2$\pm$4.7  & 54.3$\pm$4.4  & \textbf{78.8$\pm$3.6}  & \textbf{81.6$\pm$4.2} & 1.8e-4 \\
\bottomrule
\end{tabular}
\end{revised}
\end{table*}

\subsection{Trajectory Planning Performance}

\begin{revised}
TABLE~\ref{tab:results} summarizes the trajectory planning results across seen and unseen environments. We highlight five key observations. 1) \textit{C-TRAIL consistently outperforms all baselines across all metrics and scenarios}, with all improvements statistically significant at $p<0.05$. Averaged over the four simulated seen scenarios, C-TRAIL (GPT-4o) reduces ADE by 0.84 and improves SR by 17.0\,pp relative to DiLu, the strongest baseline. This confirms the effectiveness of coupling trust-weighted commonsense recall with Dirichlet-guided MCTS planning. 2) \textit{The improvements increase with scenario complexity.} On \emph{highway}, the SR gain over DiLu is $+$15.1\,pp; on \emph{intersection}, it rises to $+$20.2\,pp with the largest ADE reduction of 1.30. Complex scenarios benefit most from commonsense grounding, whereas methods without trust calibration cannot reliably exploit such semantic guidance. 3) \textit{C-TRAIL generalizes effectively from simulation to real-world traffic.} On \emph{highD} and \emph{rounD}, it maintains above 83\% SR with 15.2--17.9\,pp improvements over DiLu, demonstrating robust transfer without domain-specific tuning. The dual-trust mechanism verifies both semantic consistency and kinematic feasibility regardless of the data source. 4) \textit{C-TRAIL shows minimal degradation in unseen environments.} The average SR drop across all six unseen scenarios is only 1.7\,pp, compared with 8.4\,pp for DiLu and 7.3\,pp for GRAD. On \emph{highway}, C-TRAIL's SR decreases by merely 1.5\,pp, from 88.3\% to 86.8\%, whereas DiLu drops 9.1\,pp. This is because the trust calibration mechanism dynamically re-estimates confidence under novel configurations rather than relying on fixed priors. 5) \textit{LLM-based methods outperform classical planners, yet still trail C-TRAIL without trust calibration.} DiLu, LangMPC, and GPT-Driver achieve 2--3$\times$ higher SR than DQN and MCTS, confirming the value of semantic knowledge. However, these methods exhibit 15--20\,pp SR gaps to C-TRAIL, indicating that consuming LLM outputs without reliability assessment limits planning quality.
\end{revised}

\begin{revised}
\subsection{Ablation Study}

To assess the contribution of each component, we ablate three modules: (1)~\textit{w/o Trust}, removing trust weighting by fixing $c_{0,i,t}{=}1$ in Eq.\,(\ref{eq:trust_encoding}); (2)~\textit{w/o Dirichlet policy}, replacing the trust-based prior with a standard UCB in MCTS; and (3)~\textit{w/o Adaptive update}, using only the current-step trust score without historical smoothing. TABLE~\ref{tab:ablation_study} reports the results.

Removing the Dirichlet policy causes the largest degradation: on \emph{intersection}, SR drops from 84.2\% to 55.3\%, a loss of 28.9\,pp, confirming that an informative trust prior is essential for guiding MCTS toward semantically meaningful actions. Removing dual-trust quantification is the second most harmful: on \emph{highway}, SR falls from 88.3\% to 72.4\%, as the planner can no longer filter unreliable LLM predictions or enforce kinematic consistency. The adaptive-update ablation shows milder but consistent losses, e.g., \emph{merge} SR falls from 87.1\% to 81.0\%, indicating that temporal trust smoothing improves stability in dynamic traffic. The three components complementarily enhance C-TRAIL's performance.\looseness=-1

\begin{table*}[!t]
\centering
\begin{revised}
\small
\caption{Ablation Study of Key Modules on Planning Performance.}
\label{tab:ablation_study}
\renewcommand{\arraystretch}{1.05}
\fontsize{7.5pt}{9pt}\selectfont
\setlength{\tabcolsep}{2.5pt}
\begin{tabular}{lccc ccc ccc ccc}
\toprule
\multirow{2}{*}{\textbf{Variants}}
    & \multicolumn{3}{c}{Highway}
    & \multicolumn{3}{c}{Merge}
    & \multicolumn{3}{c}{Roundabout}
    & \multicolumn{3}{c}{Intersection} \\
\cmidrule(lr){2-4} \cmidrule(lr){5-7} \cmidrule(lr){8-10} \cmidrule(lr){11-13}
    & ADE $\downarrow$ & FDE $\downarrow$ & SR(\%) $\uparrow$
    & ADE $\downarrow$ & FDE $\downarrow$ & SR(\%) $\uparrow$
    & ADE $\downarrow$ & FDE $\downarrow$ & SR(\%) $\uparrow$
    & ADE $\downarrow$ & FDE $\downarrow$ & SR(\%) $\uparrow$ \\
\midrule
w/o Trust
                     & 1.10$\pm$0.27 & 1.45$\pm$0.31 & 72.4$\pm$5.6
                     & 1.55$\pm$0.25 & 2.34$\pm$0.42 & 70.1$\pm$4.4
                     & 2.30$\pm$0.32 & 3.40$\pm$0.52 & 68.5$\pm$4.9
                     & 2.85$\pm$0.07 & 4.75$\pm$0.09 & 65.7$\pm$3.7 \\

w/o Dirichlet policy
                     &1.45$\pm$0.30 & 1.82$\pm$0.37 & 63.5$\pm$6.2
                     & 1.85$\pm$0.26 & 2.96$\pm$0.43 & 60.2$\pm$4.7
                     & 2.78$\pm$0.36 & 3.91$\pm$0.56 & 58.7$\pm$5.4
                     & 3.12$\pm$0.11 & 5.45$\pm$0.13 & 55.3$\pm$4.3 \\

w/o Adaptive update
                     & 0.88$\pm$0.25 & 1.12$\pm$0.26 & 82.3$\pm$5.3
                     & 1.20$\pm$0.18 & 1.85$\pm$0.29 & 81.0$\pm$3.6
                     & 1.70$\pm$0.25& 2.30$\pm$0.44 & 79.8$\pm$3.5
                     & 2.05$\pm$0.05 & 3.95$\pm$0.08& 77.6$\pm$2.9 \\

\textbf{C‑TRAIL (GPT‑4o)}
                     & \textbf{0.68$\pm$0.11} & \textbf{0.80$\pm$0.09} & \textbf{88.3$\pm$1.6}
                     & \textbf{0.93$\pm$0.35} & \textbf{1.42$\pm$0.32} & \textbf{87.1$\pm$1.8}
                     & \textbf{1.38$\pm$0.25} & \textbf{1.75$\pm$0.70} & \textbf{85.2$\pm$2.2}
                     & \textbf{1.62$\pm$0.73} & \textbf{3.05$\pm$0.34} & \textbf{84.2$\pm$3.2} \\
\bottomrule
\end{tabular}
\end{revised}
\end{table*}
\end{revised}

\begin{revised}
\subsection{Safety Analysis}

\noindent\textbf{Trust Dynamics under LLM Errors.}
We track the trust score $C_t$ on the \emph{highway} scenario under two conditions: (1)~\emph{Normal} LLM predictions and (2)~\emph{Error Injection}, where 40\% of LLM-predicted relations are replaced with random incorrect relations starting at $t{=}4$ (Fig.~\ref{fig:trust_dynamics}).

\begin{figure}[!t]
\begin{revised}
    \centering
    \includegraphics[width=0.85\linewidth]{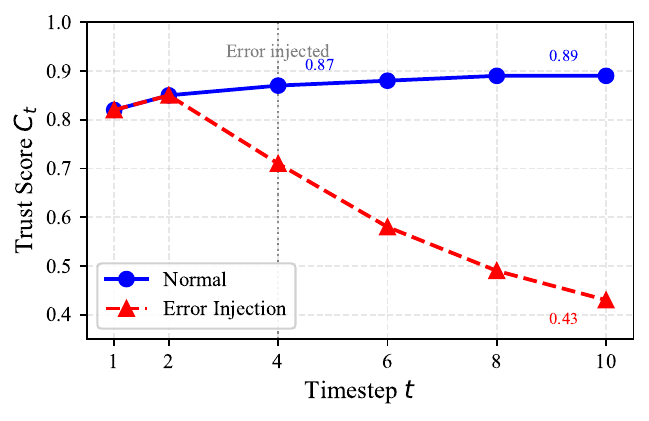}
    \caption{Trust score dynamics ($C_t$) under normal and error injection conditions on \emph{highway}. Errors are injected at $t{=}4$.}
    \label{fig:trust_dynamics}
\end{revised}
\end{figure}

Under normal conditions the trust score stabilizes at 0.89, whereas error injection at $t{=}4$ causes it to drop from 0.87 to 0.43 by $t{=}10$. The masked blending mechanism detects the discrepancy between LLM predictions and observed kinematics, triggering trust reduction. As $C_t$ falls, the Dirichlet policy flattens, redirecting MCTS toward exploration rather than unreliable LLM guidance, a self-correcting safety behavior.

\noindent\textbf{Trust Filtering Quality.}
We next measure how trust filtering improves LLM output quality. \emph{Relation Prediction Accuracy} (RPA) is the percentage of LLM-predicted pairwise relations that match ground-truth kinematic relations. TABLE~\ref{tab:trust_filtering_quality} compares three settings: no filtering, kinematic trust only ($c_{\text{kin}}$), and the full dual-trust filter ($c_{\text{llm}} \cdot c_{\text{kin}}$).

\begin{table}[!t]
\centering
\begin{revised}
\small
\caption{LLM Output Quality Before and After Trust Filtering.}
\label{tab:trust_filtering_quality}
\renewcommand{\arraystretch}{1.05}
\fontsize{7.5pt}{9pt}\selectfont
\setlength{\tabcolsep}{3pt}
\begin{tabular}{llccc}
\toprule
& & \textbf{No filtering} & $\boldsymbol{c_{\text{kin}}}$ \textbf{only} & \textbf{Dual trust (full)} \\
\midrule
\multirow{2}{*}{Highway}
 & RPA $\uparrow$ & 68.40$\pm$2.15 & 79.60$\pm$1.85 & \textbf{89.30$\pm$1.20} \\
 & SR(\%) $\uparrow$  & 74.80$\pm$2.30 & 82.70$\pm$1.95 & \textbf{88.30$\pm$1.35} \\
\midrule
\multirow{2}{*}{Merge}
 & RPA $\uparrow$ & 65.10$\pm$2.45 & 76.30$\pm$2.10 & \textbf{86.80$\pm$1.45} \\
 & SR(\%) $\uparrow$  & 71.50$\pm$2.60 & 80.30$\pm$2.20 & \textbf{87.10$\pm$1.50} \\
\midrule
\multirow{2}{*}{Roundabout}
 & RPA $\uparrow$ & 61.20$\pm$2.70 & 74.80$\pm$2.25 & \textbf{85.70$\pm$1.55} \\
 & SR(\%) $\uparrow$  & 70.30$\pm$2.85 & 78.90$\pm$2.40 & \textbf{85.20$\pm$1.70} \\
\midrule
\multirow{2}{*}{Intersection}
 & RPA $\uparrow$ & 58.70$\pm$3.10 & 71.50$\pm$2.65 & \textbf{82.40$\pm$1.90} \\
 & SR(\%) $\uparrow$  & 66.20$\pm$3.25 & 76.10$\pm$2.80 & \textbf{84.20$\pm$2.05} \\
\bottomrule
\end{tabular}
\end{revised}
\end{table}

Raw LLM output achieves only 63.4\% average RPA across four scenarios. Kinematic filtering raises this to 75.6\% by rejecting physically infeasible predictions, and the full dual-trust mechanism reaches 86.1\% by further penalizing high-entropy LLM outputs. This progressive filtering translates directly to planning: average SR improves from 70.7\% to 86.2\%, confirming that the trust mechanism serves as an effective quality gate for LLM-generated commonsense.

\noindent\textbf{Robustness to Aggressive Driving.}
We further test robustness under adversarial conditions: 30\% of NPC vehicles perform sudden cut-ins within 2 time steps at $<$15\,m distance. TABLE~\ref{tab:aggressive_npc} compares C-TRAIL, DiLu, and GRAD on \emph{highway} and \emph{roundabout}.

\begin{table}[!t]
\centering
\begin{revised}
\small
\caption{Performance Under Normal vs.\ Aggressive NPC Behavior.}
\label{tab:aggressive_npc}
\renewcommand{\arraystretch}{1.05}
\fontsize{7.5pt}{9pt}\selectfont
\setlength{\tabcolsep}{3pt}
\begin{tabular}{llccc}
\toprule
\textbf{Scenario} & \textbf{Method} & \textbf{ADE} $\downarrow$ & \textbf{FDE} $\downarrow$ & \textbf{SR(\%)} $\uparrow$ \\
\midrule
\multirow{3}{*}{\makecell[l]{Highway\\(Normal)}}
& GRAD           & 2.51$\pm$0.25 & 4.61$\pm$0.22 & 31.2$\pm$2.0 \\
& DiLu           & 1.15$\pm$0.15 & 2.81$\pm$0.17 & 73.2$\pm$1.9 \\
& C-TRAIL & \textbf{0.68$\pm$0.11} & \textbf{0.80$\pm$0.09} & \textbf{88.3$\pm$1.6} \\
\midrule
\multirow{3}{*}{\makecell[l]{Highway\\(Aggressive)}}
& GRAD           & 3.18$\pm$0.33 & 5.42$\pm$0.30 & 22.5$\pm$2.8 \\
& DiLu           & 1.72$\pm$0.22 & 3.65$\pm$0.24 & 58.4$\pm$3.1 \\
& C-TRAIL & \textbf{0.92$\pm$0.15} & \textbf{1.18$\pm$0.14} & \textbf{79.6$\pm$2.4} \\
\midrule
\multirow{3}{*}{\makecell[l]{Roundabout\\(Normal)}}
& GRAD           & 3.83$\pm$0.11 & 5.09$\pm$0.22 & 29.7$\pm$4.2 \\
& DiLu           & 2.07$\pm$0.20 & 2.94$\pm$0.33 & 68.2$\pm$3.6 \\
& C-TRAIL & \textbf{1.38$\pm$0.25} & \textbf{1.75$\pm$0.70} & \textbf{85.2$\pm$2.2} \\
\midrule
\multirow{3}{*}{\makecell[l]{Roundabout\\(Aggressive)}}
& GRAD           & 4.62$\pm$0.19 & 6.15$\pm$0.31 & 19.8$\pm$5.1 \\
& DiLu           & 2.85$\pm$0.28 & 3.92$\pm$0.41 & 52.7$\pm$4.3 \\
& C-TRAIL & \textbf{1.78$\pm$0.30} & \textbf{2.45$\pm$0.82} & \textbf{74.8$\pm$3.1} \\
\bottomrule
\end{tabular}
\end{revised}
\end{table}

Aggressive driving degrades all methods, but C-TRAIL exhibits the smallest SR drop: $-$8.7\,pp on highway and $-$10.4\,pp on roundabout, compared with $-$14.8\,pp and $-$15.5\,pp for DiLu. This resilience arises because the commonsense recall module identifies ``closing rapidly'' relations that signal potential cut-ins, and the trust mechanism lowers confidence when sudden changes invalidate assumptions, triggering wider MCTS exploration.\looseness=-1
\end{revised}

\begin{revised}
\subsection{Sensitivity Analysis}

\noindent\textbf{Trust Calibration Rate $\gamma_{\text{diri}}$.}
To investigate the effect of the trust calibration rate $\gamma_{\text{diri}}$ on planning performance, we vary its value in $\{0.1, 0.3, 0.5, 0.7, 0.9\}$ while keeping all other hyperparameters fixed. Recall from Eq.~(\ref{eq:policy_update}) that $\gamma_{\text{diri}}$ controls the balance between the historical Dirichlet prior and the newly observed trust evidence: smaller values favor temporal stability, while larger values make the policy more reactive to recent observations.

\begin{table*}[!t]
\centering
\begin{revised}
\small
\caption{Sensitivity of Trust Calibration Rate $\gamma_{\text{diri}}$ on Planning Performance.}
\label{tab:gamma_diri_ablation}
\renewcommand{\arraystretch}{1.05}
\fontsize{7.5pt}{9pt}\selectfont
\setlength{\tabcolsep}{2.5pt}
\begin{tabular}{lccc ccc ccc ccc}
\toprule
\multirow{2}{*}{$\gamma_{\text{diri}}$}
    & \multicolumn{3}{c}{Highway}
    & \multicolumn{3}{c}{Merge}
    & \multicolumn{3}{c}{Roundabout}
    & \multicolumn{3}{c}{Intersection} \\
\cmidrule(lr){2-4} \cmidrule(lr){5-7} \cmidrule(lr){8-10} \cmidrule(lr){11-13}
    & ADE $\downarrow$ & FDE $\downarrow$ & SR(\%) $\uparrow$
    & ADE $\downarrow$ & FDE $\downarrow$ & SR(\%) $\uparrow$
    & ADE $\downarrow$ & FDE $\downarrow$ & SR(\%) $\uparrow$
    & ADE $\downarrow$ & FDE $\downarrow$ & SR(\%) $\uparrow$ \\
\midrule
0.1 & 0.82$\pm$0.14 & 0.98$\pm$0.12 & 84.5$\pm$2.3
    & 1.15$\pm$0.39 & 1.72$\pm$0.37 & 83.2$\pm$2.4
    & 1.62$\pm$0.28 & 2.08$\pm$0.74 & 81.3$\pm$2.8
    & 1.95$\pm$0.78 & 3.52$\pm$0.40 & 79.8$\pm$3.6 \\

\textbf{0.3} & \textbf{0.68$\pm$0.11} & \textbf{0.80$\pm$0.09} & \textbf{88.3$\pm$1.6}
    & \textbf{0.93$\pm$0.35} & \textbf{1.42$\pm$0.32} & \textbf{87.1$\pm$1.8}
    & \textbf{1.38$\pm$0.25} & \textbf{1.75$\pm$0.70} & \textbf{85.2$\pm$2.2}
    & \textbf{1.62$\pm$0.73} & \textbf{3.05$\pm$0.34} & \textbf{84.2$\pm$3.2} \\

0.5 & 0.75$\pm$0.13 & 0.91$\pm$0.11 & 86.1$\pm$2.0
    & 1.05$\pm$0.37 & 1.58$\pm$0.35 & 84.8$\pm$2.1
    & 1.51$\pm$0.27 & 1.93$\pm$0.72 & 83.0$\pm$2.5
    & 1.78$\pm$0.75 & 3.28$\pm$0.37 & 81.7$\pm$3.4 \\

0.7 & 0.91$\pm$0.16 & 1.08$\pm$0.14 & 82.7$\pm$2.5
    & 1.28$\pm$0.41 & 1.89$\pm$0.39 & 80.5$\pm$2.8
    & 1.78$\pm$0.31 & 2.35$\pm$0.78 & 78.4$\pm$3.1
    & 2.15$\pm$0.80 & 3.85$\pm$0.43 & 76.3$\pm$3.8 \\

0.9 & 1.12$\pm$0.19 & 1.35$\pm$0.18 & 78.2$\pm$3.1
    & 1.52$\pm$0.44 & 2.24$\pm$0.42 & 75.8$\pm$3.3
    & 2.08$\pm$0.35 & 2.78$\pm$0.82 & 73.6$\pm$3.7
    & 2.58$\pm$0.85 & 4.32$\pm$0.48 & 70.5$\pm$4.2 \\
\bottomrule
\end{tabular}
\end{revised}
\end{table*}

As shown in TABLE~\ref{tab:gamma_diri_ablation}, the default setting $\gamma_{\text{diri}}=0.3$ yields the best overall performance across all four scenarios. When $\gamma_{\text{diri}}$ is too small, e.g., 0.1, the trust prior updates too slowly, causing the Dirichlet policy to rely excessively on stale historical evidence and respond sluggishly to changing traffic dynamics. Conversely, when $\gamma_{\text{diri}}$ is too large, e.g., 0.7 or 0.9, the policy becomes overly reactive to noisy single-step observations, leading to unstable action distributions and degraded planning quality. For instance, at $\gamma_{\text{diri}}=0.9$, ADE increases by 64.7\% and SR drops by 10.1\% on \emph{highway} compared to the optimal setting. The moderate value $\gamma_{\text{diri}}=0.3$ strikes the best balance between adaptability and stability, enabling the trust calibration mechanism to smoothly incorporate new evidence while retaining informative historical priors.

\noindent\textbf{Traffic Density and Lane Variations.}
To assess the robustness of C-TRAIL to diverse traffic, we vary the lane number and vehicle density across all four scenarios using the configurations described in Section~\ref{sec:data_collection}.

\begin{table}[!t]
\centering
\begin{revised}
\small
\caption{Robustness to Lane Count and Vehicle Density.}
\label{tab:robustness}
\renewcommand{\arraystretch}{1.05}
\fontsize{7.5pt}{9pt}\selectfont
\setlength{\tabcolsep}{3pt}
\begin{tabular}{ll cc cc}
\toprule
& & \multicolumn{2}{c}{Density 2.0} & \multicolumn{2}{c}{Density 3.0} \\
\cmidrule(lr){3-4} \cmidrule(lr){5-6}
Scenario & Lanes & ADE $\downarrow$ & FDE $\downarrow$ & ADE $\downarrow$ & FDE $\downarrow$ \\
\midrule
\multirow{2}{*}{Highway}
& 4 & 0.68$\pm$0.11 & 0.80$\pm$0.09 & 0.74$\pm$0.12 & 0.88$\pm$0.11 \\
& 5 & 0.71$\pm$0.12 & 0.84$\pm$0.11 & 0.78$\pm$0.13 & 0.92$\pm$0.12 \\
\midrule
\multirow{2}{*}{Merge}
& 4 & 0.93$\pm$0.35 & 1.42$\pm$0.32 & 1.05$\pm$0.42 & 1.57$\pm$0.37 \\
& 5 & 0.97$\pm$0.36 & 1.48$\pm$0.33 & 1.11$\pm$0.38 & 1.64$\pm$0.35 \\
\midrule
Roundabout & 2 & 1.38$\pm$0.25 & 1.75$\pm$0.70 & 1.45$\pm$0.28 & 1.87$\pm$0.75 \\
\midrule
Intersection & 2 & 1.62$\pm$0.73 & 3.05$\pm$0.34 & 1.72$\pm$0.78 & 3.23$\pm$0.40 \\
\bottomrule
\end{tabular}
\end{revised}
\end{table}

As shown in TABLE~\ref{tab:robustness}, C-TRAIL maintains consistently low ADE and FDE across all configurations. Higher vehicle density, e.g., from 2.0 to 3.0, causes a modest increase in planning error, while varying lane number has limited impact, especially in \emph{highway} and \emph{merge} scenarios. This suggests that C-TRAIL is more sensitive to traffic density than road width, consistent with the intuition that denser traffic increases planning difficulty. Overall, C‑TRAIL demonstrates strong generalization under increasingly complex conditions.
\end{revised}

\begin{revised}
\subsection{Computational Efficiency}

TABLE~\ref{tab:efficiency_comparison} compares the computational efficiency of LLM-based methods across simulated and real-world scenarios. We focus on LLM-based approaches since non-LLM baselines such as DQN, MCTS, and GRAD are orders of magnitude faster at $<$1\,s by design but achieve substantially lower planning accuracy, as shown in TABLE~\ref{tab:results}. The bottom section of the table further decomposes C-TRAIL's inference time into three components: 1) \textbf{\textit{LLM API Call}} (prompt construction and commonsense querying), 2) \textbf{\textit{MCTS Planning}} ($K{=}50$ simulations with Dirichlet-guided selection), and 3) \textbf{\textit{Trust Update}} (trust recalibration).

\begin{table}[!t]
\centering
\begin{revised}
\small
\setlength{\tabcolsep}{1.5pt}
\caption{Computational Efficiency of LLM-Based Methods.}
\label{tab:efficiency_comparison}
\renewcommand{\arraystretch}{1.05}
\fontsize{6.5pt}{8pt}\selectfont
\begin{tabular}{lll cccc}
\toprule
 & & & \textbf{Highway} & \textbf{Roundab.} & \textbf{highD} & \textbf{rounD} \\
\midrule
\multirow{2}{*}{LMTraj}
 & Infer (s) & & 1.56$\pm$0.14 & 1.73$\pm$0.20 & 1.61$\pm$0.16 & 1.78$\pm$0.22 \\
 & Mem (MB) & & 276$\pm$4 & 295$\pm$4 & 278$\pm$5 & 298$\pm$5 \\
\cmidrule{1-7}
\multirow{2}{*}{DiLu}
 & Infer (s) & & 15.39$\pm$0.50 & 17.81$\pm$0.37 & 15.72$\pm$0.48 & 18.05$\pm$0.42 \\
 & Mem (MB) & & 613$\pm$19 & 662$\pm$20 & 621$\pm$20 & 670$\pm$21 \\
\cmidrule{1-7}
\multirow{2}{*}{LangMPC}
 & Infer (s) & & 12.35$\pm$0.39 & 13.42$\pm$0.36 & 12.52$\pm$0.38 & 13.58$\pm$0.40 \\
 & Mem (MB) & & 485$\pm$14 & 510$\pm$15 & 490$\pm$14 & 515$\pm$15 \\
\cmidrule{1-7}
\multirow{2}{*}{GPT-Dr}
 & Infer (s) & & 16.72$\pm$0.60 & 17.83$\pm$0.54 & 16.90$\pm$0.58 & 18.02$\pm$0.56 \\
 & Mem (MB) & & 578$\pm$13 & 602$\pm$13 & 582$\pm$14 & 609$\pm$14 \\
\midrule
\multirow{5}{*}{C-TRAIL}
 & \multirow{4}{*}{Infer (s)} & LLM Call & 13.42$\pm$0.82 & 13.78$\pm$0.88 & 13.55$\pm$0.84 & 13.85$\pm$0.90 \\
 & & MCTS & 3.78$\pm$0.20 & 3.90$\pm$0.21 & 3.80$\pm$0.20 & 3.95$\pm$0.22 \\
 & & Trust & 0.80$\pm$0.07 & 0.82$\pm$0.07 & 0.80$\pm$0.07 & 0.82$\pm$0.07 \\
 & & \textbf{Overall} & \textbf{18.00$\pm$1.06} & \textbf{18.50$\pm$1.13} & \textbf{18.15$\pm$1.08} & \textbf{18.62$\pm$1.15} \\
\cmidrule{2-7}
 & Mem (MB) & & \textbf{536$\pm$11} & \textbf{560$\pm$12} & \textbf{540$\pm$11} & \textbf{563$\pm$13} \\
\bottomrule
\end{tabular}
\end{revised}
\end{table}

C-TRAIL (GPT-4o) incurs an inference time of 18.00--18.62\,s with peak memory of 536--563\,MB. Among LLM-based methods, C-TRAIL offers comparable or lower memory usage than DiLu at 613--670\,MB and GPT-Driver at 578--608\,MB, while delivering substantially better planning accuracy. Although LMTraj is faster at 1.56--1.78\,s due to its single-pass fine-tuned architecture, it achieves significantly lower SR across all scenarios as shown in TABLE~\ref{tab:results}. The LLM API call dominates C-TRAIL's inference latency, accounting for approximately 74.6\% of the total time on \emph{highway} and 74.4\% on \emph{rounD}. MCTS planning accounts for 21.0\%, while trust update is lightweight at about 4.4\%. Latency increases slightly in complex scenarios such as \emph{roundabout} and \emph{rounD} due to more surrounding vehicles requiring additional LLM queries and MCTS rollouts. The latency on real-world datasets closely matches simulated counterparts, confirming that the computational overhead is driven by the inference pipeline rather than dataset characteristics.
\end{revised}

\begin{revised}
\subsection{Case Study}

We visualize representative planned trajectories to illustrate how C-TRAIL succeeds and where it fails across four driving scenarios.

\begin{figure}[!t]
    \centering
    \subfigure[Highway]{%
        \includegraphics[width=\linewidth]{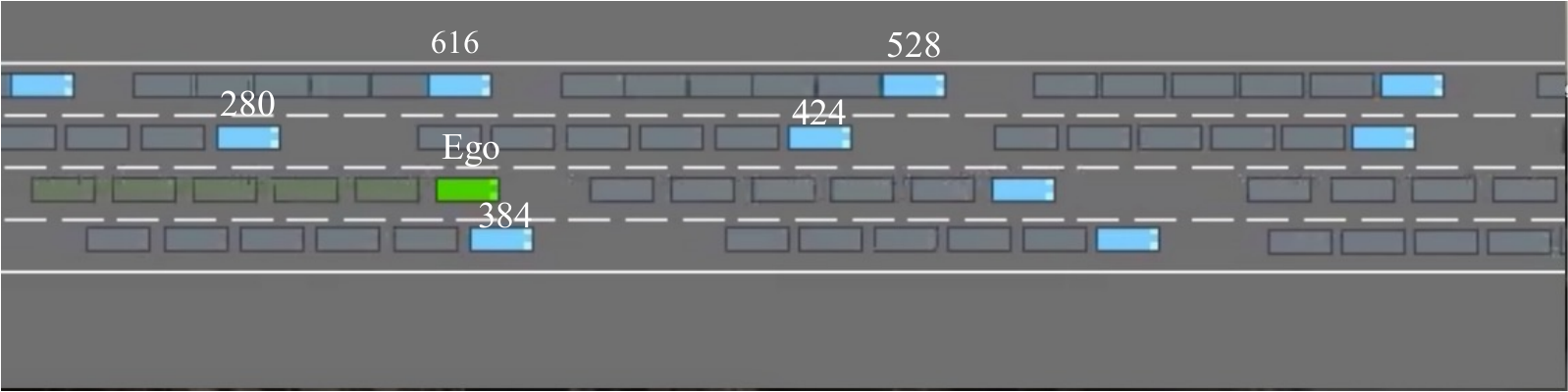}
        \label{fig:ego-highway-trajectory}
    }\\[0.2em]
    \subfigure[Merge]{%
        \includegraphics[width=\linewidth]{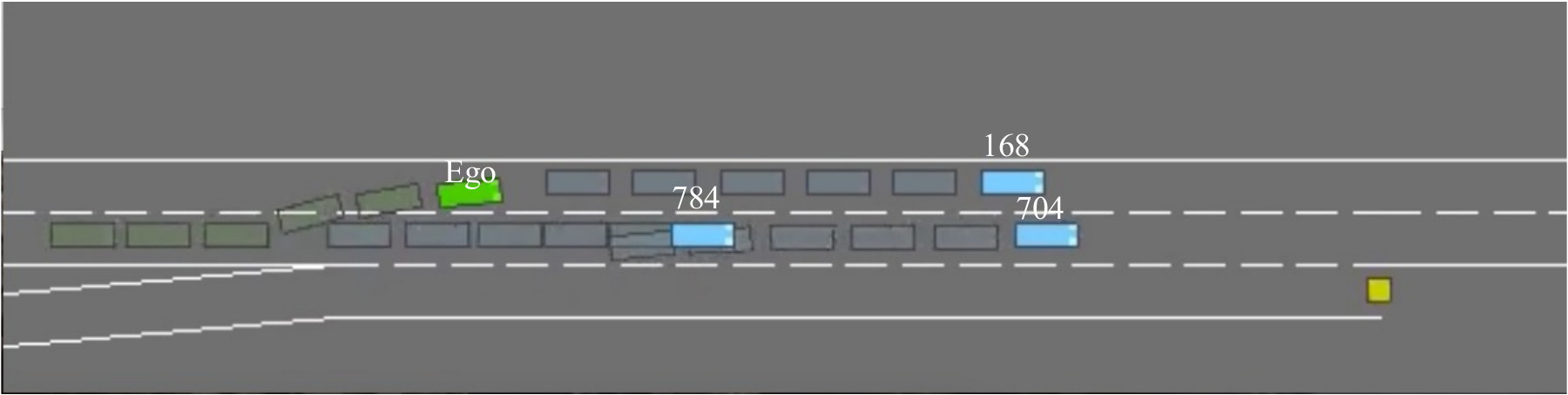}
        \label{fig:ego-merge-trajectory}
    }\\[0.2em]
    \subfigure[Roundabout]{%
        \includegraphics[width=0.47\linewidth]{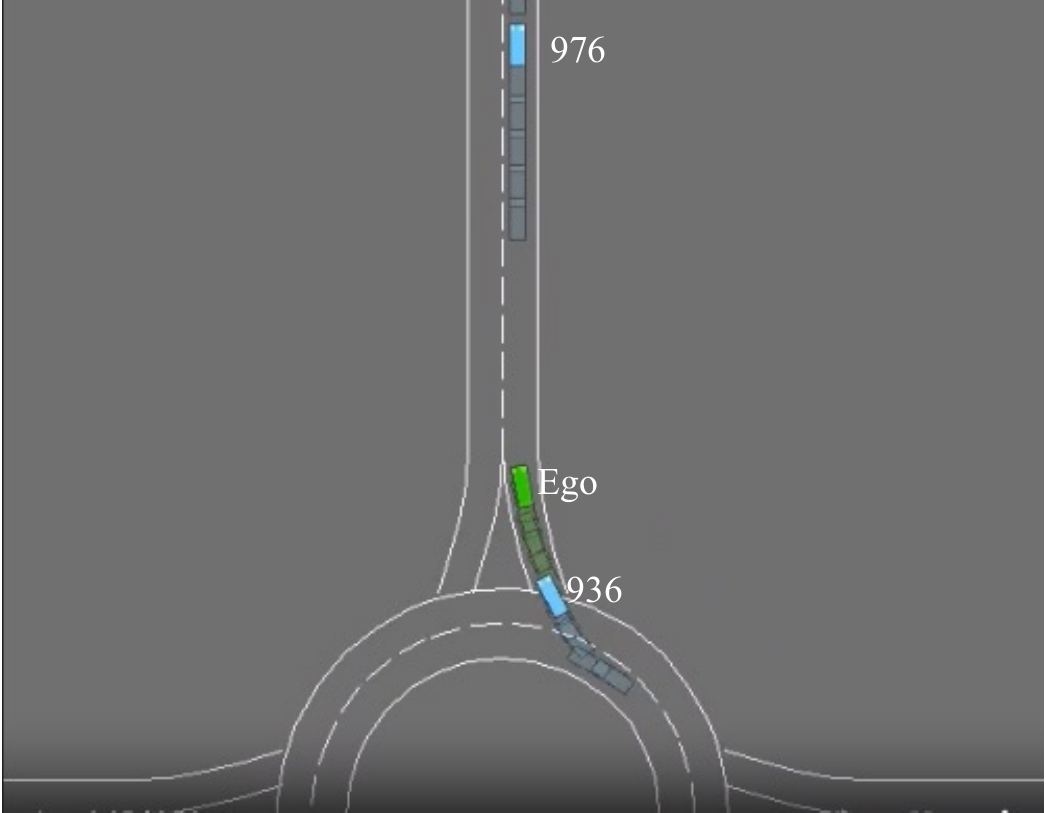}
        \label{fig:ego-roundabout-trajectory}
    }
    \subfigure[Intersection]{%
        \includegraphics[width=0.46\linewidth]{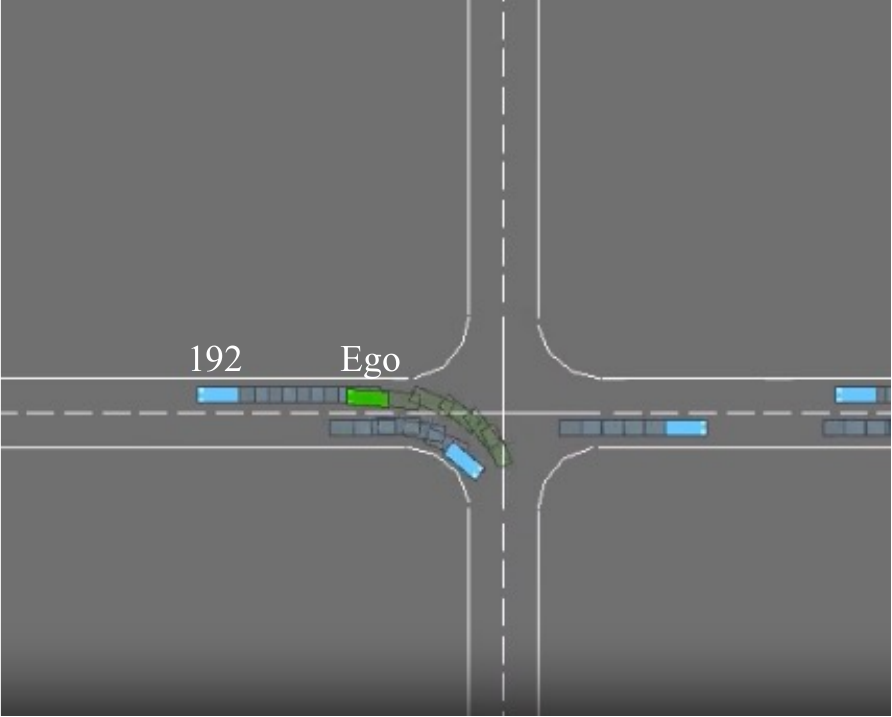}
        \label{fig:ego-intersection-trajectory}
    }
    \caption{Successfully executed planned trajectories across four scenarios. The green box denotes the ego vehicle, which follows the planned 5-step trajectory.
Blue boxes represent surrounding vehicles controlled by the simulator.}
    \label{fig:Trajectories_Visualization}
\end{figure}

\noindent\textbf{Successful Trajectories.} As shown in Fig.~\ref{fig:Trajectories_Visualization}, the ego vehicle successfully completes all four tasks following trajectories generated by C-TRAIL. The planned paths maintain safe distances from surrounding vehicles and adapt to scenario-specific challenges, including lane-keeping on highways, gap selection during merging, yielding in roundabouts, and crossing at intersections.

\begin{figure}[!t]
    \centering
    \subfigure[Highway]{%
        \includegraphics[width=\linewidth]{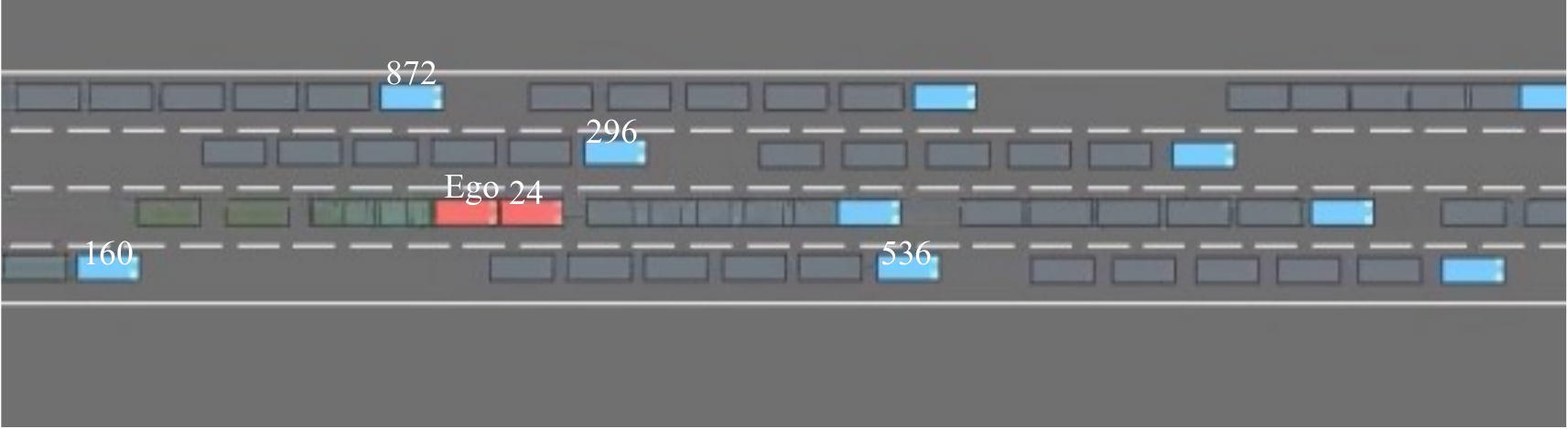}
        \label{fig:failed-ego-highway-trajectory}
    }\\[0.2em]
    \subfigure[Merge]{%
        \includegraphics[width=\linewidth]{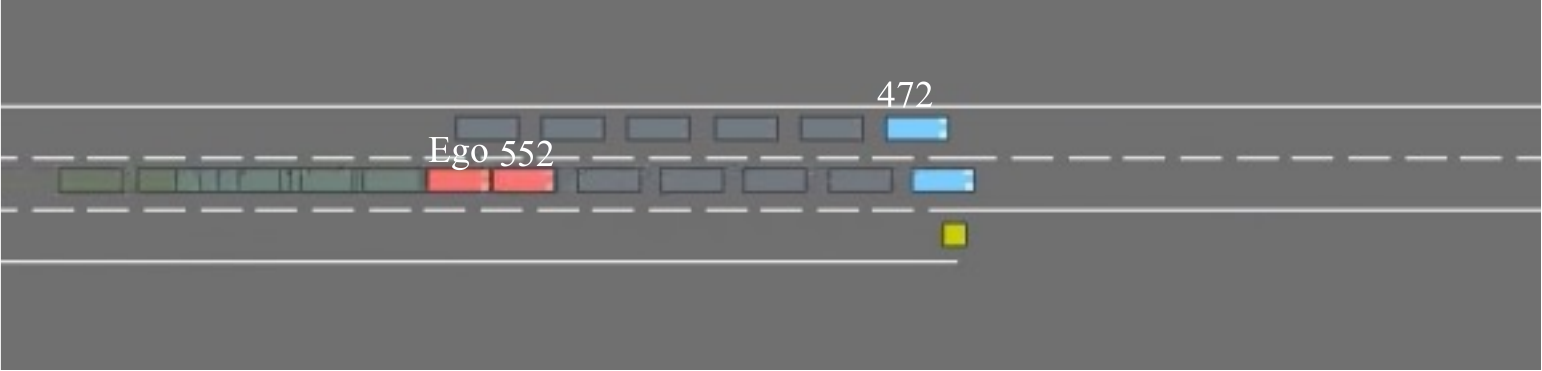}
        \label{fig:failed-ego-merge-trajectory}
    }\\[0.2em]
    \subfigure[Roundabout]{%
        \includegraphics[width=0.45\linewidth]{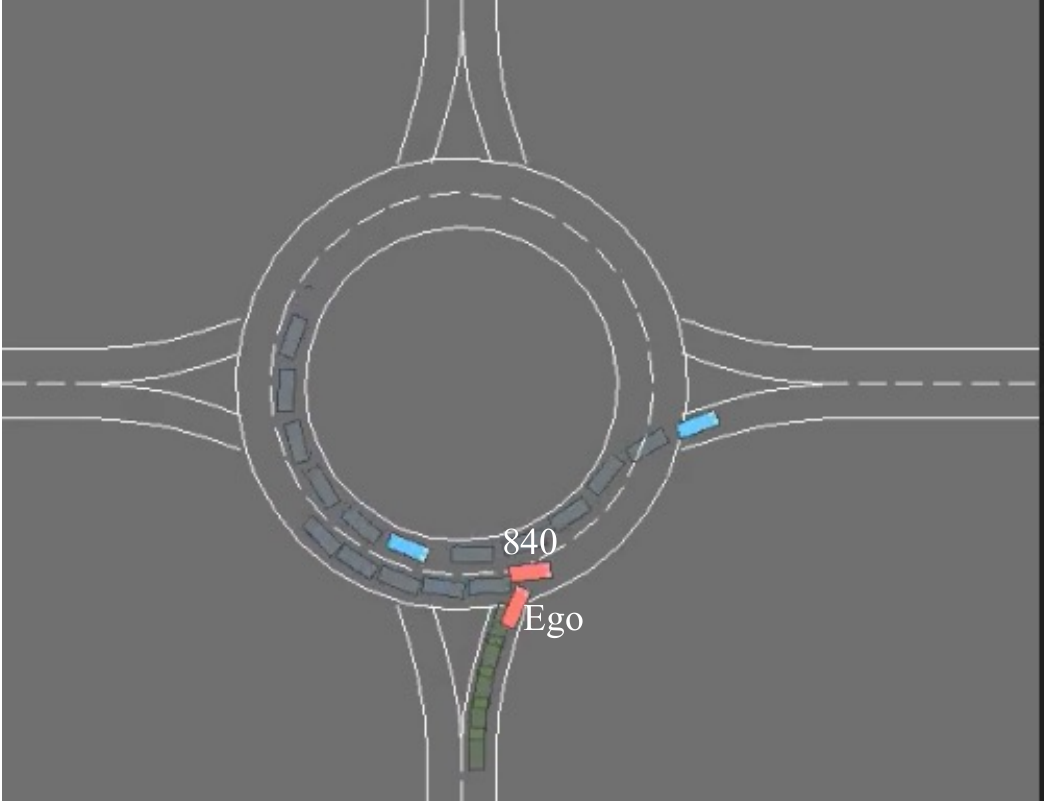}
        \label{fig:failed-ego-roundabout-trajectory}
    }
    \subfigure[Intersection]{%
        \includegraphics[width=0.48\linewidth]{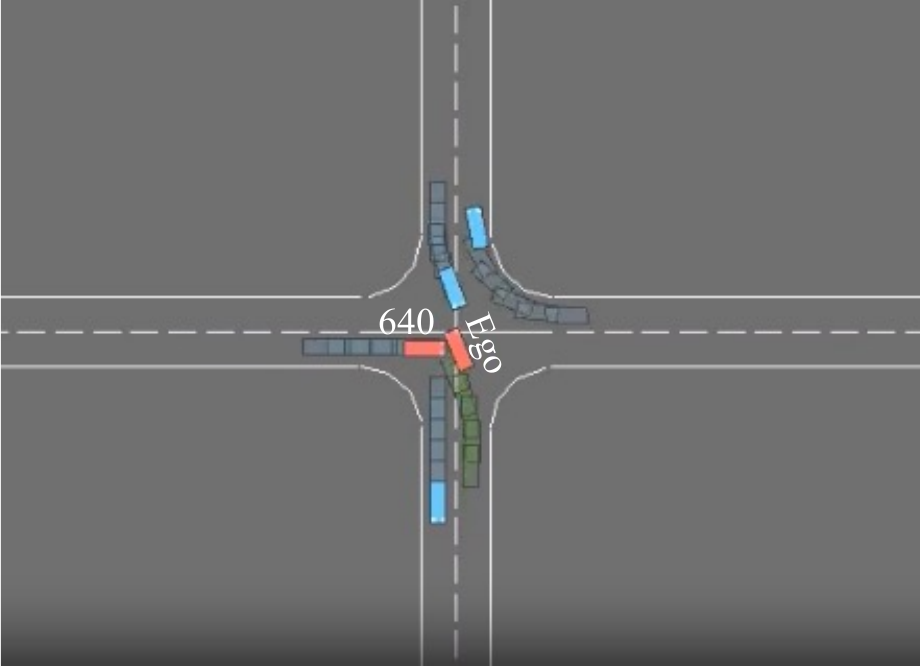}
        \label{fig:failed-ego-intersection-trajectory}
    }
    \caption{Failed planned trajectories resulting in collisions, illustrating three failure types: policy error~(a), model error~(c), and mapping error~(d).}
    \label{fig:Failed_Trajectories_Visualization}
\end{figure}

\noindent\textbf{Failure Case.} Fig.~\ref{fig:Failed_Trajectories_Visualization} presents representative failure cases, which fall into three categories.
\emph{Policy errors} occur when the Dirichlet-based policy assigns comparable probabilities to conflicting actions, causing Q-value oscillations in MCTS; e.g., the planner selects ``acceleration'' despite vehicle~24 being directly ahead (Fig.~\ref{fig:failed-ego-highway-trajectory}).
\emph{Model errors} arise from incomplete commonsense representations; for instance, an approaching vehicle~840 is filtered during trust recall, leading the planner to assume the lane is clear (Fig.~\ref{fig:failed-ego-roundabout-trajectory}).
\emph{Mapping errors} occur when a correct LLM intent (e.g., ``turn left'') is incorrectly grounded to a different simulator action such as IDLE (Fig.~\ref{fig:failed-ego-intersection-trajectory}).

\begin{table}[!t]
\centering
\begin{revised}
\small
\caption{Failure Breakdown (\%) in Seen Environments.}
\label{tab:failure_breakdown}
\renewcommand{\arraystretch}{1.05}
\fontsize{7.5pt}{9pt}\selectfont
\setlength{\tabcolsep}{3pt}
\begin{tabular}{lccc}
\toprule
\textbf{Scenario} & \textbf{Policy Error} & \textbf{Model Error} & \textbf{Mapping Error} \\
\midrule
Highway      & 52.30$\pm$3.12 & 28.60$\pm$2.71 & 19.10$\pm$2.38 \\
highD        & 50.80$\pm$3.42 & 30.40$\pm$2.93 & 18.80$\pm$2.51 \\
\midrule
Roundabout   & 45.10$\pm$3.24 & 35.80$\pm$3.05 & 19.10$\pm$2.33 \\
rounD        & 42.60$\pm$3.48 & 38.50$\pm$3.15 & 18.90$\pm$2.58 \\
\midrule
\textbf{Average} & \textbf{47.70$\pm$3.32} & \textbf{33.33$\pm$2.96} & \textbf{19.00$\pm$2.45} \\
\bottomrule
\end{tabular}
\end{revised}
\end{table}

TABLE~\ref{tab:failure_breakdown} quantifies the failure distribution across simulated and real-world scenarios. Policy errors dominate at 47.7\%, particularly in high-speed highway settings where time-critical decisions amplify Q-value oscillations. Model errors are more prevalent in roundabout scenarios at 35.8--38.5\% due to complex spatial interactions. Sim-to-real pairs exhibit similar distributions, confirming that the dominant failure modes transfer across domains. These findings motivate targeted mitigations: ensemble LLM queries with majority voting for policy errors, conservative trust thresholds with rule-based fallback for model errors, and verified action grounding for mapping errors.
\end{revised}

\section{Related Work}

We categorize related work in trajectory planning for autonomous driving according to three stages in C-TRAIL: Recall, Plan, and Update. TABLE~\ref{relatedwork} summarizes representative methods across these three stages.

\subsection{Recall: LLM-based Commonsense Recall}
Recent studies~\cite{li2018humanlike, hou2024my} have highlighted the potential of LLMs to retrieve contextual knowledge for autonomous driving tasks. DiLu~\cite{wen2023dilu} proposes a reasoning-reflection framework that incorporates commonsense into decision-making, and LC-LLM~\cite{peng2024lc} reformulates lane-change prediction as a language modeling task. Other works~\cite{arora2022ask, wang2023voyager, brohan2023can} explore prompt engineering and structured inputs to improve recall quality, while domain-specific adaptations such as TrafficSafetyGPT~\cite{zheng2023trafficsafetygpt} fine-tune LLMs for transportation safety. Mahmud et al.~\cite{mahmud2025integrating} further survey the broader integration of LLMs into intelligent transportation systems. Despite these advances, the recalled knowledge is generally consumed without assessing its reliability, leaving downstream planning modules exposed to potentially noisy or hallucinated guidance.

\subsection{Plan: Trajectory Generation and Planning}
Classical planning methods such as AlphaGo~\cite{silver2017mastering} and MuZero~\cite{schrittwieser2020mastering} demonstrate the effectiveness of MCTS for long-horizon decision-making. In driving contexts, Wen et al.~\cite{wen2024monte} and MBAPPE~\cite{chekroun2024mbappe} incorporate cost functions or learned dynamics into MCTS, and IGP2~\cite{albrecht2021interpretable} combines inverse planning with MCTS for goal inference. These methods, however, rely on handcrafted objectives and lack semantic-level guidance. LLM-based planners such as Code-as-Policies~\cite{liang2023code} and Huang et al.~\cite{huang2022language} attempt to generate plans directly but struggle with generalization and real-time demands. Emerging efforts bridge the two paradigms: Sun et al.~\cite{sun2024prompt} elicit action priors from LLMs for few-shot planning, and Du et al.~\cite{du2023guiding} pretrain RL policies using LLM-generated knowledge, yet neither provides a mechanism to regulate the influence of LLM-derived guidance during the search process.

\subsection{Update: Real-Time Adaptation}
Static planners tend to degrade as the environment shifts beyond the training distribution~\cite{liu2019trajectory, li2022learning}. Voyager~\cite{wang2023voyager} addresses this through continual LLM-driven exploration, and TidyBot~\cite{wu2023tidybot} updates task plans based on user preferences; however, both rely on offline adaptation and cannot respond at the frequency required by real-time driving. Parameter-efficient tuning methods such as dLoRA~\cite{wu2024dlora} and incremental frameworks like IETA~\cite{han2023ieta} support continual model updates but depend on large-scale streaming feedback. Lightweight alternatives such as Agent-Pro~\cite{zhang2024agent} refine policies via reflection-optimization with minimal supervision, yet still lack a principled mechanism to continuously evaluate the reliability of commonsense knowledge that informs planning decisions.

\begin{revised}
In conclusion, existing methods typically address only one or two of the three stages. As shown in TABLE~\ref{relatedwork}, DiLu~\cite{wen2023dilu} designs a dedicated reasoning-reflection pipeline for Recall ($\checkmark$), but its planning remains rule-driven without trajectory search ($\Delta$) and its reflection operates offline rather than continuously ($\Delta$). In contrast, Wen et al.~\cite{wen2024monte} provide a rigorous MCTS-based planner ($\checkmark$) while leaving knowledge retrieval and online adaptation entirely unaddressed ($\times$). TidyBot~\cite{wu2023tidybot}, though covering both Recall and Plan ($\checkmark$), is limited to preference-based offline adaptation for Update ($\Delta$). Meanwhile, Agent-Pro~\cite{zhang2024agent} focuses solely on policy refinement through reflection-optimization ($\checkmark$ for Update), without incorporating any recall or planning mechanism. None of the existing methods jointly addresses all three stages with dedicated mechanisms. This observation motivates the proposed C-TRAIL, which jointly addresses all three stages through a unified trust-aware closed-loop mechanism that continuously calibrates commonsense reliability during both trajectory planning and real-time adaptation.

\begin{table}[!t]
\centering
\begin{revised}
\caption{Comparison of representative methods across the Recall, Plan, and Update stages in trajectory planning. $\checkmark$: the method explicitly designs a dedicated module or mechanism for the stage; $\Delta$: the stage is partially addressed as a byproduct of the main design, without a dedicated or complete mechanism; $\times$: the stage is not addressed.}
\label{relatedwork}
\begin{tabular}{lccc}
\toprule
\textbf{Method} & \textbf{Recall} & \textbf{Plan} & \textbf{Update} \\
\midrule
DiLu~\cite{wen2023dilu} & $\checkmark$ & $\Delta$ & $\Delta$ \\
LC-LLM~\cite{peng2024lc} & $\checkmark$ & $\checkmark$ & $\times$ \\
Voyager~\cite{wang2023voyager} & $\checkmark$ & $\Delta$ & $\checkmark$ \\
TrafficSafetyGPT~\cite{zheng2023trafficsafetygpt} & $\checkmark$ & $\times$ & $\times$ \\
Wen et al.~\cite{wen2024monte} & $\times$ & $\checkmark$ & $\times$ \\
MBAPPE~\cite{chekroun2024mbappe} & $\times$ & $\checkmark$ & $\times$ \\
FAIR Diplomacy~\cite{meta2022human} & $\checkmark$ & $\checkmark$ & $\times$ \\
Sun et al.~\cite{sun2024prompt} & $\checkmark$ & $\checkmark$ & $\times$ \\
TidyBot~\cite{wu2023tidybot} & $\checkmark$ & $\checkmark$ & $\Delta$ \\
Agent-Pro~\cite{zhang2024agent} & $\times$ & $\times$ & $\checkmark$ \\
Code-as-Policies~\cite{liang2023code} & $\times$ & $\checkmark$ & $\times$ \\
Huang et al.~\cite{huang2022language} & $\checkmark$ & $\checkmark$ & $\times$ \\
Du et al.~\cite{du2023guiding} & $\checkmark$ & $\checkmark$ & $\times$ \\
dLoRA~\cite{wu2024dlora} & $\times$ & $\times$ & $\checkmark$ \\
IETA~\cite{han2023ieta} & $\times$ & $\times$ & $\checkmark$ \\
\bottomrule
\end{tabular}
\end{revised}
\end{table}
\end{revised}

\section{Conclusion}
\label{sec:conclusion}

This work presented C-TRAIL, a trust-aware framework that integrates LLM commonsense into trajectory planning through a closed-loop Recall--Plan--Update cycle. By quantifying LLM reliability via a dual-trust mechanism and injecting trust-weighted guidance into MCTS through a Dirichlet policy, C-TRAIL converts unreliable LLM outputs into calibrated planning priors. \rev{Experiments on four Highway-env scenarios and two real-world levelXData datasets show consistent improvements over baselines, reducing ADE by 40.2\%, FDE by 51.7\%, and improving SR by 16.9\,pp.} \rev{Ablation studies confirm that the dual-trust mechanism is especially critical in unseen environments, where removing trust assessment degrades SR by up to 18.5\,pp.}

\begin{revised}
\noindent\textbf{Limitations and Future Work.}
Although the trust mechanism empirically detects LLM errors and degrades gracefully, formal safety verification for LLM-integrated planners remains an open challenge. Future work includes extending C-TRAIL to multi-agent interaction-aware planning, incorporating multimodal inputs for richer commonsense grounding, and validating on high-fidelity simulators such as CARLA~\cite{dosovitskiy2017carla} and nuPlan~\cite{caesar2021nuplan} that offer continuous vehicle dynamics beyond the kinematic model used here.
\end{revised}

\section*{Acknowledgment}
This work has received funding from the European Union’s Horizon Europe research and innovation programme under the Marie Skłodowska-Curie grant agreement No. 101126636.

\bibliographystyle{IEEEtran}

\begin{thebibliography}{10}
\providecommand{\url}[1]{#1}
\csname url@samestyle\endcsname
\providecommand{\newblock}{\relax}
\providecommand{\bibinfo}[2]{#2}
\providecommand{\BIBentrySTDinterwordspacing}{\spaceskip=0pt\relax}
\providecommand{\BIBentryALTinterwordstretchfactor}{4}
\providecommand{\BIBentryALTinterwordspacing}{\spaceskip=\fontdimen2\font plus
\BIBentryALTinterwordstretchfactor\fontdimen3\font minus
  \fontdimen4\font\relax}
\providecommand{\BIBforeignlanguage}[2]{{%
\expandafter\ifx\csname l@#1\endcsname\relax
\typeout{** WARNING: IEEEtran.bst: No hyphenation pattern has been}%
\typeout{** loaded for the language `#1'. Using the pattern for}%
\typeout{** the default language instead.}%
\else
\language=\csname l@#1\endcsname
\fi
#2}}
\providecommand{\BIBdecl}{\relax}
\BIBdecl

\bibitem{hu2023planning}
Y.~Hu, J.~Yang, L.~Chen, K.~Li, C.~Sima, X.~Zhu, S.~Chai, S.~Du, T.~Lin,
  W.~Wang \emph{et~al.}, ``Planning-oriented autonomous driving,'' in
  \emph{Proceedings of the IEEE/CVF conference on computer vision and pattern
  recognition}, 2023, pp. 17\,853--17\,862.

\bibitem{chen2024planning}
G.~Chen, M.~Hua, W.~Liu, J.~Wang, S.~Song, C.~Liu, L.~Yang, S.~Liao, and
  X.~Xia, ``Planning and tracking control of full drive-by-wire electric
  vehicles in unstructured scenario,'' \emph{Proceedings of the Institution of
  Mechanical Engineers}, vol. 238, no.~13, pp. 3941--3956, 2024.

\bibitem{li2023trajectory}
H.~Li, P.~Chen, G.~Yu, B.~Zhou, Y.~Li, and Y.~Liao, ``Trajectory planning for
  autonomous driving in unstructured scenarios based on deep learning and
  quadratic optimization,'' \emph{IEEE Transactions on Vehicular Technology},
  vol.~73, no.~4, pp. 4886--4903, 2023.

\bibitem{groves2014framework}
W.~Groves, E.~Nunes, and M.~Gini, ``A framework for predicting trajectories
  using global and local information,'' in \emph{Proceedings of the 11th ACM
  Conference on Computing Frontiers}, 2014, pp. 1--10.

\bibitem{xie2023cognition}
S.~Xie, J.~Li, and J.~Wang, ``A cognition-inspired trajectory prediction method
  for vehicles in interactive scenarios,'' \emph{IET Intelligent Transport
  Systems}, vol.~17, no.~8, pp. 1544--1559, 2023.

\bibitem{zhang2024predicting}
Z.~Zhang, A.~Li, A.~Lim, and M.~Chen, ``Predicting long-term human behaviors in
  discrete representations via physics-guided diffusion,'' \emph{arXiv preprint
  arXiv:2405.19528}, 2024.

\bibitem{codevilla2019exploring}
F.~Codevilla, E.~Santana, A.~M. L{\'o}pez, and A.~Gaidon, ``Exploring the
  limitations of behavior cloning for autonomous driving,'' in
  \emph{Proceedings of the IEEE/CVF international conference on computer
  vision}, 2019, pp. 9329--9338.

\bibitem{teng2023motion}
S.~Teng, X.~Hu, P.~Deng, B.~Li, Y.~Li, Y.~Ai, D.~Yang, L.~Li, Z.~Xuanyuan,
  F.~Zhu \emph{et~al.}, ``Motion planning for autonomous driving: The state of
  the art and future perspectives,'' \emph{IEEE Transactions on Intelligent
  Vehicles}, vol.~8, no.~6, pp. 3692--3711, 2023.

\bibitem{epstein2017cognitive}
R.~A. Epstein, E.~Z. Patai, J.~B. Julian, and H.~J. Spiers, ``The cognitive map
  in humans: spatial navigation and beyond,'' \emph{Nature neuroscience},
  vol.~20, no.~11, pp. 1504--1513, 2017.

\bibitem{son2024replay}
J.-Y. Son, M.-L. Vives, A.~Bhandari, and O.~FeldmanHall, ``Replay shapes
  abstract cognitive maps for efficient social navigation,'' \emph{Nature Human
  Behaviour}, vol.~8, no.~11, pp. 2156--2167, 2024.

\bibitem{song2024trial}
Y.~Song, D.~Yin, X.~Yue, J.~Huang, S.~Li, and B.~Y. Lin, ``Trial and error:
  Exploration-based trajectory optimization for llm agents,'' \emph{arXiv
  preprint arXiv:2403.02502}, 2024.

\bibitem{felemban2024imotion}
A.~Felemban, E.~M. Bakr, X.~Shen, J.~Ding, A.~Mohamed, and M.~Elhoseiny,
  ``imotion-llm: Motion prediction instruction tuning,'' \emph{arXiv preprint
  arXiv:2406.06211}, 2024.

\bibitem{li2025synergy}
X.~Li, Y.~Sun, J.~Lin, L.~Li, T.~Feng, and S.~Yin, ``The synergy of seeing and
  saying: Revolutionary advances in multi-modality medical vision-language
  large models,'' \emph{Artificial Intelligence Science and Engineering},
  vol.~1, no.~2, pp. 79--97, 2025.

\bibitem{tang2025dybooster}
H.~Tang, X.~Sun, S.~Wu, Z.~Cui, G.~Xu, and Q.~Li, ``{DyBooster}: Leveraging
  large language model as booster for dynamic recommendation,'' \emph{Expert
  Systems with Applications}, vol. 286, p. 128080, 2025.

\bibitem{peng2023check}
B.~Peng, M.~Galley, P.~He, H.~Cheng, Y.~Xie, Y.~Hu, Q.~Huang, L.~Liden, Z.~Yu,
  W.~Chen \emph{et~al.}, ``Check your facts and try again: Improving large
  language models with external knowledge and automated feedback,'' \emph{arXiv
  preprint arXiv:2302.12813}, 2023.

\bibitem{huang2025survey}
L.~Huang, W.~Yu, W.~Ma, W.~Zhong, Z.~Feng, H.~Wang, Q.~Chen, W.~Peng, X.~Feng,
  B.~Qin \emph{et~al.}, ``A survey on hallucination in large language models:
  Principles, taxonomy, challenges, and open questions,'' \emph{ACM
  Transactions on Information Systems}, vol.~43, no.~2, pp. 1--55, 2025.

\bibitem{wang2024q}
C.~Wang, Y.~Deng, Z.~Lyu, L.~Zeng, J.~He, S.~Yan, and B.~An, ``Q*: Improving
  multi-step reasoning for llms with deliberative planning,'' \emph{arXiv
  preprint arXiv:2406.14283}, 2024.

\bibitem{li2024embodied}
M.~Li, S.~Zhao, Q.~Wang, K.~Wang, Y.~Zhou, S.~Srivastava, C.~Gokmen, T.~Lee,
  E.~L. Li, R.~Zhang \emph{et~al.}, ``Embodied agent interface: Benchmarking
  llms for embodied decision making,'' \emph{Advances in Neural Information
  Processing Systems}, vol.~37, pp. 100\,428--100\,534, 2024.

\bibitem{sun2023adaplanner}
H.~Sun, Y.~Zhuang, L.~Kong, B.~Dai, and C.~Zhang, ``Adaplanner: Adaptive
  planning from feedback with language models,'' \emph{Advances in neural
  information processing systems}, vol.~36, pp. 58\,202--58\,245, 2023.

\bibitem{pan2023automatically}
L.~Pan, M.~Saxon, W.~Xu, D.~Nathani, X.~Wang, and W.~Y. Wang, ``Automatically
  correcting large language models: Surveying the landscape of diverse
  self-correction strategies,'' \emph{arXiv preprint arXiv:2308.03188}, 2023.

\bibitem{mahmud2025integrating}
D.~Mahmud, H.~Hajmohamed, S.~Almentheri, S.~Alqaydi, L.~Aldhaheri, R.~A.
  Khalil, and N.~Saeed, ``Integrating llms with its: Recent advances,
  potentials, challenges, and future directions,'' \emph{IEEE Transactions on
  Intelligent Transportation Systems}, 2025.

\bibitem{zheng2023trafficsafetygpt}
O.~Zheng, M.~Abdel-Aty, D.~Wang, C.~Wang, and S.~Ding, ``Trafficsafetygpt:
  Tuning a pre-trained large language model to a domain-specific expert in
  transportation safety,'' \emph{arXiv preprint arXiv:2307.15311}, 2023.

\bibitem{wang2023survey}
C.~Wang, X.~Liu, Y.~Yue, X.~Tang, T.~Zhang, C.~Jiayang, Y.~Yao, W.~Gao, X.~Hu,
  Z.~Qi \emph{et~al.}, ``Survey on factuality in large language models:
  Knowledge, retrieval and domain-specificity,'' \emph{arXiv preprint
  arXiv:2310.07521}, 2023.

\bibitem{brohan2023can}
A.~Brohan, Y.~Chebotar, C.~Finn, K.~Hausman, A.~Herzog, D.~Ho, J.~Ibarz,
  A.~Irpan, E.~Jang, R.~Julian \emph{et~al.}, ``Do as i can, not as i say:
  Grounding language in robotic affordances,'' in \emph{Conference on robot
  learning}.\hskip 1em plus 0.5em minus 0.4em\relax PMLR, 2023, pp. 287--318.

\bibitem{arora2022ask}
S.~Arora, A.~Narayan, M.~F. Chen, L.~Orr, N.~Guha, K.~Bhatia, I.~Chami,
  F.~Sala, and C.~R{\'e}, ``Ask me anything: A simple strategy for prompting
  language models,'' \emph{arXiv preprint arXiv:2210.02441}, 2022.

\bibitem{huang2022language}
W.~Huang, P.~Abbeel, D.~Pathak, and I.~Mordatch, ``Language models as zero-shot
  planners: Extracting actionable knowledge for embodied agents,'' in
  \emph{International conference on machine learning}.\hskip 1em plus 0.5em
  minus 0.4em\relax PMLR, 2022, pp. 9118--9147.

\bibitem{silver2017mastering}
D.~Silver, J.~Schrittwieser, K.~Simonyan, I.~Antonoglou, A.~Huang, A.~Guez,
  T.~Hubert, L.~Baker, M.~Lai, A.~Bolton \emph{et~al.}, ``Mastering the game of
  go without human knowledge,'' \emph{nature}, vol. 550, no. 7676, pp.
  354--359, 2017.

\bibitem{guan2023leveraging}
L.~Guan, K.~Valmeekam, S.~Sreedharan, and S.~Kambhampati, ``Leveraging
  pre-trained large language models to construct and utilize world models for
  model-based task planning,'' \emph{Advances in Neural Information Processing
  Systems}, vol.~36, pp. 79\,081--79\,094, 2023.

\bibitem{song2023llm}
C.~H. Song, J.~Wu, C.~Washington, B.~M. Sadler, W.-L. Chao, and Y.~Su,
  ``Llm-planner: Few-shot grounded planning for embodied agents with large
  language models,'' in \emph{Proceedings of the IEEE/CVF international
  conference on computer vision}, 2023, pp. 2998--3009.

\bibitem{liang2023code}
J.~Liang, W.~Huang, F.~Xia, P.~Xu, K.~Hausman, B.~Ichter, P.~Florence, and
  A.~Zeng, ``Code as policies: Language model programs for embodied control,''
  in \emph{2023 IEEE International Conference on Robotics and Automation
  (ICRA)}.\hskip 1em plus 0.5em minus 0.4em\relax IEEE, 2023, pp. 9493--9500.

\bibitem{schrittwieser2020mastering}
J.~Schrittwieser, I.~Antonoglou, T.~Hubert, K.~Simonyan, L.~Sifre, S.~Schmitt,
  A.~Guez, E.~Lockhart, D.~Hassabis, T.~Graepel \emph{et~al.}, ``Mastering
  atari, go, chess and shogi by planning with a learned model,'' \emph{Nature},
  vol. 588, no. 7839, pp. 604--609, 2020.

\bibitem{meta2022human}
M.~F. A. R. D.~T. (FAIR)†, A.~Bakhtin, N.~Brown, E.~Dinan, G.~Farina,
  C.~Flaherty, D.~Fried, A.~Goff, J.~Gray, H.~Hu \emph{et~al.}, ``Human-level
  play in the game of diplomacy by combining language models with strategic
  reasoning,'' \emph{Science}, vol. 378, no. 6624, pp. 1067--1074, 2022.

\bibitem{liu2024evolution}
F.~Liu, X.~Tong, M.~Yuan, X.~Lin, F.~Luo, Z.~Wang, Z.~Lu, and Q.~Zhang,
  ``Evolution of heuristics: Towards efficient automatic algorithm design using
  large language model,'' \emph{arXiv preprint arXiv:2401.02051}, 2024.

\bibitem{sun2024prompt}
J.~Sun, Q.~Zhang, Y.~Duan, X.~Jiang, C.~Cheng, and R.~Xu, ``Prompt, plan,
  perform: Llm-based humanoid control via quantized imitation learning,'' in
  \emph{2024 IEEE International Conference on Robotics and Automation
  (ICRA)}.\hskip 1em plus 0.5em minus 0.4em\relax IEEE, 2024, pp.
  16\,236--16\,242.

\bibitem{du2023guiding}
Y.~Du, O.~Watkins, Z.~Wang, C.~Colas, T.~Darrell, P.~Abbeel, A.~Gupta, and
  J.~Andreas, ``Guiding pretraining in reinforcement learning with large
  language models,'' in \emph{International Conference on Machine
  Learning}.\hskip 1em plus 0.5em minus 0.4em\relax PMLR, 2023, pp. 8657--8677.

\bibitem{wang2023voyager}
G.~Wang, Y.~Xie, Y.~Jiang, A.~Mandlekar, C.~Xiao, Y.~Zhu, L.~Fan, and
  A.~Anandkumar, ``Voyager: An open-ended embodied agent with large language
  models,'' \emph{arXiv preprint arXiv:2305.16291}, 2023.

\bibitem{wu2023tidybot}
J.~Wu, R.~Antonova, A.~Kan, M.~Lepert, A.~Zeng, S.~Song, J.~Bohg,
  S.~Rusinkiewicz, and T.~Funkhouser, ``Tidybot: Personalized robot assistance
  with large language models,'' \emph{Autonomous Robots}, vol.~47, no.~8, pp.
  1087--1102, 2023.

\bibitem{wang2024large}
J.~Wang, S.~Kai, L.~Luo, W.~Wei, Y.~Hu, A.~W.-C. Liew, S.~Pan, and B.~Yin,
  ``Large language models-guided dynamic adaptation for temporal knowledge
  graph reasoning,'' \emph{Advances in Neural Information Processing Systems},
  vol.~37, pp. 8384--8410, 2024.

\bibitem{wu2024dlora}
B.~Wu, R.~Zhu, Z.~Zhang, P.~Sun, X.~Liu, and X.~Jin, ``$\{$dLoRA$\}$:
  Dynamically orchestrating requests and adapters for $\{$LoRA$\}$$\{$LLM$\}$
  serving,'' in \emph{18th USENIX Symposium on Operating Systems Design and
  Implementation (OSDI 24)}, 2024, pp. 911--927.

\bibitem{tang2025model}
H.~Tang, S.~Wu, Z.~Cui, Y.~Li, G.~Xu, and Q.~Li, ``Model-agnostic dual-side
  online fairness learning for dynamic recommendation,'' \emph{IEEE
  Transactions on Knowledge and Data Engineering}, vol.~37, no.~5, pp.
  2727--2742, 2025.

\bibitem{han2023ieta}
J.~Han, H.~Liu, S.~Liu, X.~Chen, N.~Tan, H.~Chai, and H.~Xiong, ``ieta: A
  robust and scalable incremental learning framework for time-of-arrival
  estimation,'' in \emph{Proceedings of the 29th ACM SIGKDD Conference on
  Knowledge Discovery and Data Mining}, 2023, pp. 4100--4111.

\bibitem{zhang2024agent}
W.~Zhang, K.~Tang, H.~Wu, M.~Wang, Y.~Shen, G.~Hou, Z.~Tan, P.~Li, Y.~Zhuang,
  and W.~Lu, ``Agent-pro: Learning to evolve via policy-level reflection and
  optimization,'' \emph{arXiv preprint arXiv:2402.17574}, 2024.

\bibitem{tian2023learning}
Q.~Tian, K.~Kuang, F.~Liu, and B.~Wang, ``Learning from good trajectories in
  offline multi-agent reinforcement learning,'' in \emph{Proceedings of the
  AAAI Conference on Artificial Intelligence}, vol.~37, no.~10, 2023, pp.
  11\,672--11\,680.

\bibitem{li2025optimal}
C.~Li and W.~Wang, ``Optimal impulse control and impulse game for
  continuous-time deterministic systems: A review,'' \emph{Artificial
  Intelligence Science and Engineering}, vol.~1, no.~3, pp. 208--219, 2025.

\bibitem{silver2018general}
D.~Silver, T.~Hubert, J.~Schrittwieser, I.~Antonoglou, M.~Lai, A.~Guez,
  M.~Lanctot, L.~Sifre, D.~Kumaran, T.~Graepel \emph{et~al.}, ``A general
  reinforcement learning algorithm that masters chess, shogi, and go through
  self-play,'' \emph{Science}, vol. 362, no. 6419, pp. 1140--1144, 2018.

\bibitem{highway-env}
E.~Leurent, ``An environment for autonomous driving decision-making,''
  \url{https://github.com/eleurent/highway-env}, 2018.

\bibitem{wen2023dilu}
L.~Wen, D.~Fu, X.~Li, X.~Cai, T.~Ma, P.~Cai, M.~Dou, B.~Shi, L.~He, and
  Y.~Qiao, ``Dilu: A knowledge-driven approach to autonomous driving with large
  language models,'' \emph{arXiv preprint arXiv:2309.16292}, 2023.

\bibitem{cui2024survey}
C.~Cui, Y.~Ma, X.~Cao, W.~Ye, Y.~Zhou, K.~Liang, J.~Chen, J.~Lu, Z.~Yang, K.-D.
  Liao \emph{et~al.}, ``A survey on multimodal large language models for
  autonomous driving,'' in \emph{Proceedings of the IEEE/CVF Winter Conference
  on Applications of Computer Vision}, 2024, pp. 958--979.

\bibitem{xi2022graph}
Z.~Xi and G.~Sukthankar, ``A graph representation for autonomous driving,'' in
  \emph{The 36th Conference on Neural Information Processing Systems Workshop},
  vol.~7, no.~8, 2022, p.~9.

\bibitem{highDdataset}
R.~Krajewski, J.~Bock, L.~Kloeker, and L.~Eckstein, ``The highd dataset: A
  drone dataset of naturalistic vehicle trajectories on german highways for
  validation of highly automated driving systems,'' in \emph{2018 21st
  International Conference on Intelligent Transportation Systems (ITSC)}, 2018,
  pp. 2118--2125.

\bibitem{rounDdataset}
R.~Krajewski, T.~Moers, J.~Bock, L.~Vater, and L.~Eckstein, ``The round
  dataset: A drone dataset of road user trajectories at roundabouts in
  germany,'' in \emph{2020 IEEE 23rd International Conference on Intelligent
  Transportation Systems (ITSC)}, 2020, pp. 1--6.

\bibitem{mnih2015human}
V.~Mnih, K.~Kavukcuoglu, D.~Silver, A.~A. Rusu, J.~Veness, M.~G. Bellemare,
  A.~Graves, M.~Riedmiller, A.~K. Fidjeland, G.~Ostrovski \emph{et~al.},
  ``Human-level control through deep reinforcement learning,'' \emph{Nature},
  vol. 518, no. 7540, pp. 529--533, 2015.

\bibitem{song2025inverse}
W.~Song and S.~Tong, ``Inverse reinforcement learning optimal control for
  takagi-sugeno fuzzy systems,'' \emph{Artificial Intelligence Science and
  Engineering}, vol.~1, no.~2, pp. 134--146, 2025.

\bibitem{coulom2006efficient}
R.~Coulom, ``Efficient selectivity and backup operators in monte-carlo tree
  search,'' in \emph{International Conference on Computers and Games}.\hskip
  1em plus 0.5em minus 0.4em\relax Springer, 2006, pp. 72--83.

\bibitem{liu2025mhgin}
H.~Liu, T.~Li, Y.~He, K.~Torp, Y.~Li, and C.~S. Jensen, ``{MH-GIN}: Multi-scale
  heterogeneous graph-based imputation network for {AIS} data,''
  \emph{Proceedings of the VLDB Endowment}, vol.~19, no.~2, pp. 170--182, 2025.

\bibitem{bae2024can}
I.~Bae, J.~Lee, and H.-G. Jeon, ``Can language beat numerical regression?
  language-based multimodal trajectory prediction,'' in \emph{Proceedings of
  the IEEE/CVF Conference on Computer Vision and Pattern Recognition}, 2024,
  pp. 753--766.

\bibitem{sha2023languagempc}
H.~Sha, Y.~Mu, Y.~Jiang, L.~Chen, C.~Xu, P.~Luo, S.~E. Li, M.~Tomizuka,
  W.~Zhan, and M.~Ding, ``Languagempc: Large language models as decision makers
  for autonomous driving,'' \emph{arXiv preprint arXiv:2310.03026}, 2023.

\bibitem{mao2023gpt}
J.~Mao, Y.~Qian, H.~Zhao, and Y.~Wang, ``Gpt-driver: Learning to drive with
  gpt,'' \emph{arXiv preprint arXiv:2310.01415}, 2023.

\bibitem{li2018humanlike}
L.~Li, K.~Ota, and M.~Dong, ``Humanlike driving: Empirical decision-making
  system for autonomous vehicles,'' \emph{IEEE Transactions on Vehicular
  Technology}, vol.~67, no.~8, pp. 6814--6823, 2018.

\bibitem{hou2024my}
Y.~Hou, H.~Tamoto, and H.~Miyashita, ``" my agent understands me better":
  Integrating dynamic human-like memory recall and consolidation in llm-based
  agents,'' in \emph{Extended Abstracts of the CHI Conference on Human Factors
  in Computing Systems}, 2024, pp. 1--7.

\bibitem{peng2024lc}
M.~Peng, X.~Guo, X.~Chen, M.~Zhu, K.~Chen, X.~Wang, Y.~Wang \emph{et~al.},
  ``Lc-llm: Explainable lane-change intention and trajectory predictions with
  large language models,'' \emph{arXiv preprint arXiv:2403.18344}, 2024.

\bibitem{wen2024monte}
Q.~Wen, Z.~Gong, L.~Zhou, and Z.~Zhang, ``Monte carlo tree search for behavior
  planning in autonomous driving,'' in \emph{2024 IEEE International Symposium
  on Safety Security Rescue Robotics (SSRR)}.\hskip 1em plus 0.5em minus
  0.4em\relax IEEE, 2024.

\bibitem{chekroun2024mbappe}
R.~Chekroun, T.~Gilles, M.~Toromanoff, S.~Hornauer, and F.~Moutarde, ``Mbappe:
  Mcts-built-around prediction for planning explicitly,'' in \emph{2024 IEEE
  Intelligent Vehicles Symposium (IV)}.\hskip 1em plus 0.5em minus 0.4em\relax
  IEEE, 2024, pp. 2062--2069.

\bibitem{albrecht2021interpretable}
S.~V. Albrecht, C.~Brewitt, J.~Wilhelm, B.~Gyevnar, F.~Eiras, M.~Dobre, and
  S.~Ramamoorthy, ``Interpretable goal-based prediction and planning for
  autonomous driving,'' in \emph{2021 IEEE International Conference on Robotics
  and Automation (ICRA)}.\hskip 1em plus 0.5em minus 0.4em\relax IEEE, 2021,
  pp. 1043--1049.

\bibitem{liu2019trajectory}
X.~Liu, M.~Zhang, and E.~Rogers, ``Trajectory tracking control for autonomous
  underwater vehicles based on fuzzy re-planning of a local desired
  trajectory,'' \emph{IEEE Transactions on Vehicular Technology}, vol.~68,
  no.~12, pp. 11\,657--11\,667, 2019.

\bibitem{li2022learning}
Z.~Li, P.~Zhao, C.~Jiang, W.~Huang, and H.~Liang, ``A learning-based model
  predictive trajectory planning controller for automated driving in
  unstructured dynamic environments,'' \emph{IEEE Transactions on Vehicular
  Technology}, vol.~71, no.~6, pp. 5944--5959, 2022.

\bibitem{dosovitskiy2017carla}
A.~Dosovitskiy, G.~Ros, F.~Codevilla, A.~Lopez, and V.~Koltun, ``{CARLA}: An
  open urban driving simulator,'' in \emph{Conference on Robot Learning
  (CoRL)}, 2017, pp. 1--16.

\bibitem{caesar2021nuplan}
H.~Caesar, J.~Kabzan, K.~S. Tan, W.~K. Fong, E.~Wolber, A.~Lang, L.~Fletcher,
  O.~Beijbom, and S.~Omari, ``nu{P}lan: A closed-loop {ML}-based planning
  benchmark for autonomous vehicles,'' \emph{arXiv preprint arXiv:2106.11810},
  2021.

\end{thebibliography}

\vskip -1\baselineskip plus -1fil

\begin{IEEEbiography}[{\includegraphics[width=1in,height=1.25in,clip,keepaspectratio]{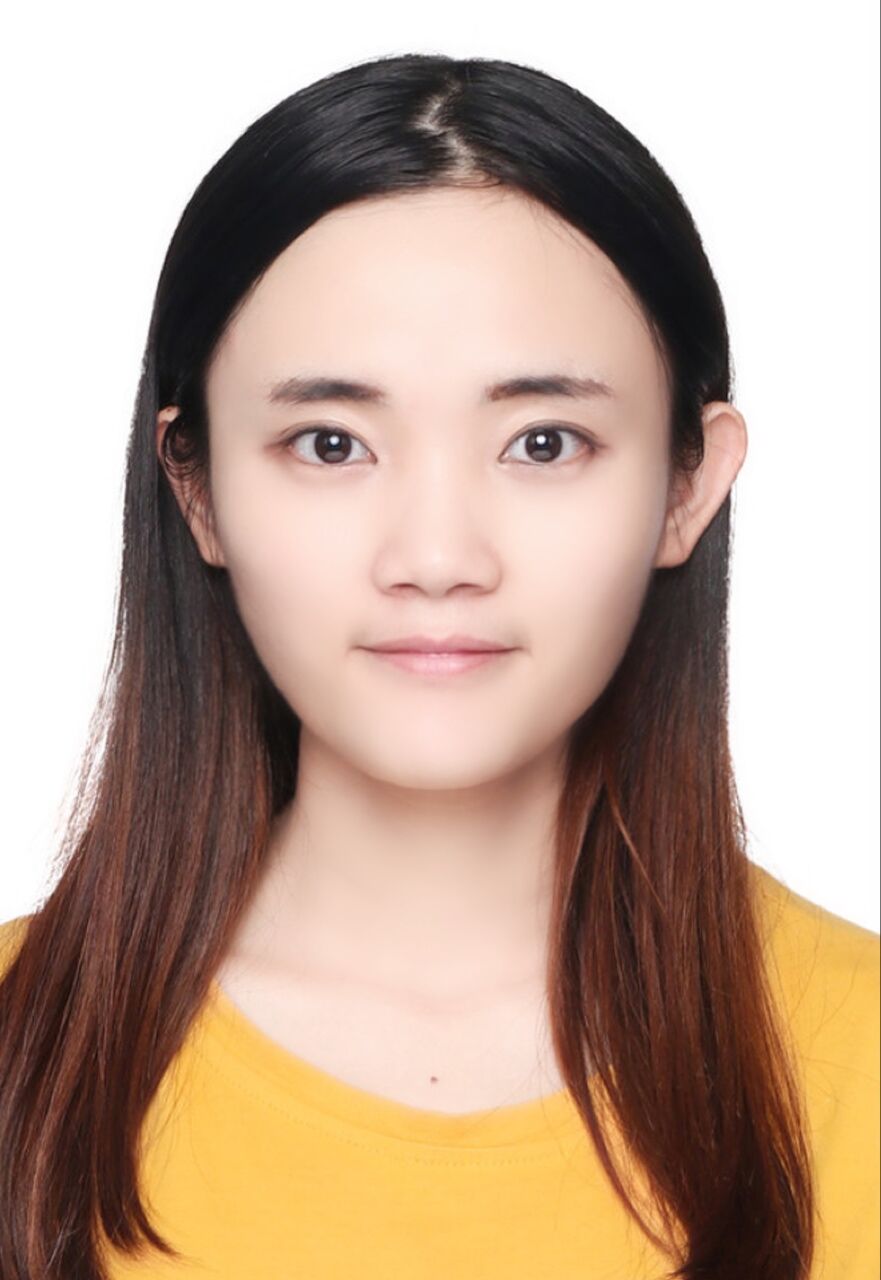}}]{Zhihong Cui} is currently a postdoc at the University of Oslo, Norway. She has received her PhD from Shandong University in 2024 supervised by Prof. Shijun Liu.  During her PhD, she worked as a Joint PhD student at the University of Technology of Sydney under the supervision of Prof. Guandong Xu. Her research interests include autonomous driving, data mining, recommendation system. Her work has been published in some of the most prestigious journals, such as IEEE TKDE, Information Sciences, IEEE TCSS, et al.
\end{IEEEbiography}

\vskip -1\baselineskip plus -1fil
\begin{IEEEbiography}[{\includegraphics[width=1in,height=1.25in,clip,keepaspectratio]{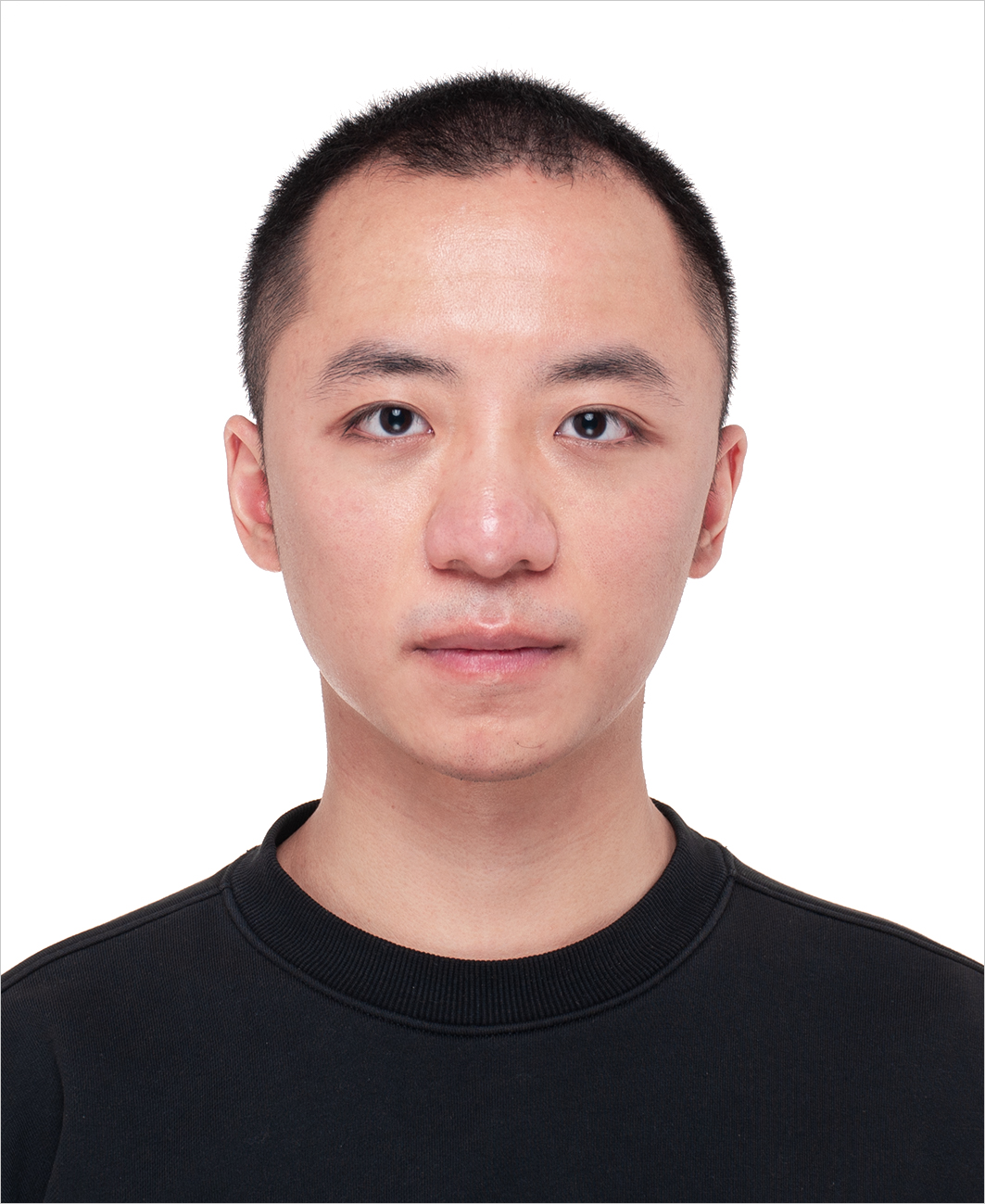}}]{Haoran Tang} is currently a Ph.D. student at the Hong Kong Polytechnic University and the University of Technology Sydney, jointly supervised by Prof. Qing Li and Prof. Guandong Xu. He received BSc, MSc degrees from Chongqing University. His research interests include dynamic recommendation, interaction network, and data mining. He has published several research papers in some top-tier conferences and journals, such as SIGIR, TOIS, TKDE, and KDD.
\end{IEEEbiography}

\vskip -1\baselineskip plus -1fil
\begin{IEEEbiography}[{\includegraphics[width=1in,height=1.25in,clip,keepaspectratio]{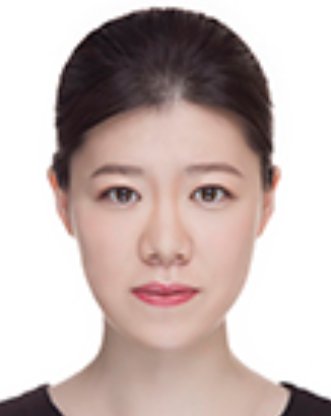}}] {Tianyi Li} (Member, IEEE) received the Ph.D. degree from Aalborg University, Denmark, in 2022. She is an Assistant Professor at the Department of Computer Science, Aalborg University. She received ICDE 2022 Best Paper Award.  She serves as an Associate Editor for IEEE Network and IEEE Transactions on Intelligent Vehicles. Her research concerns spatio-temporal data, intelligent transportation, machine learning, time series, and database technology.
\end{IEEEbiography}
\vskip -1\baselineskip plus -1fil

\begin{IEEEbiography}
[{\includegraphics[width=1in,height=1.25in,clip,keepaspectratio]{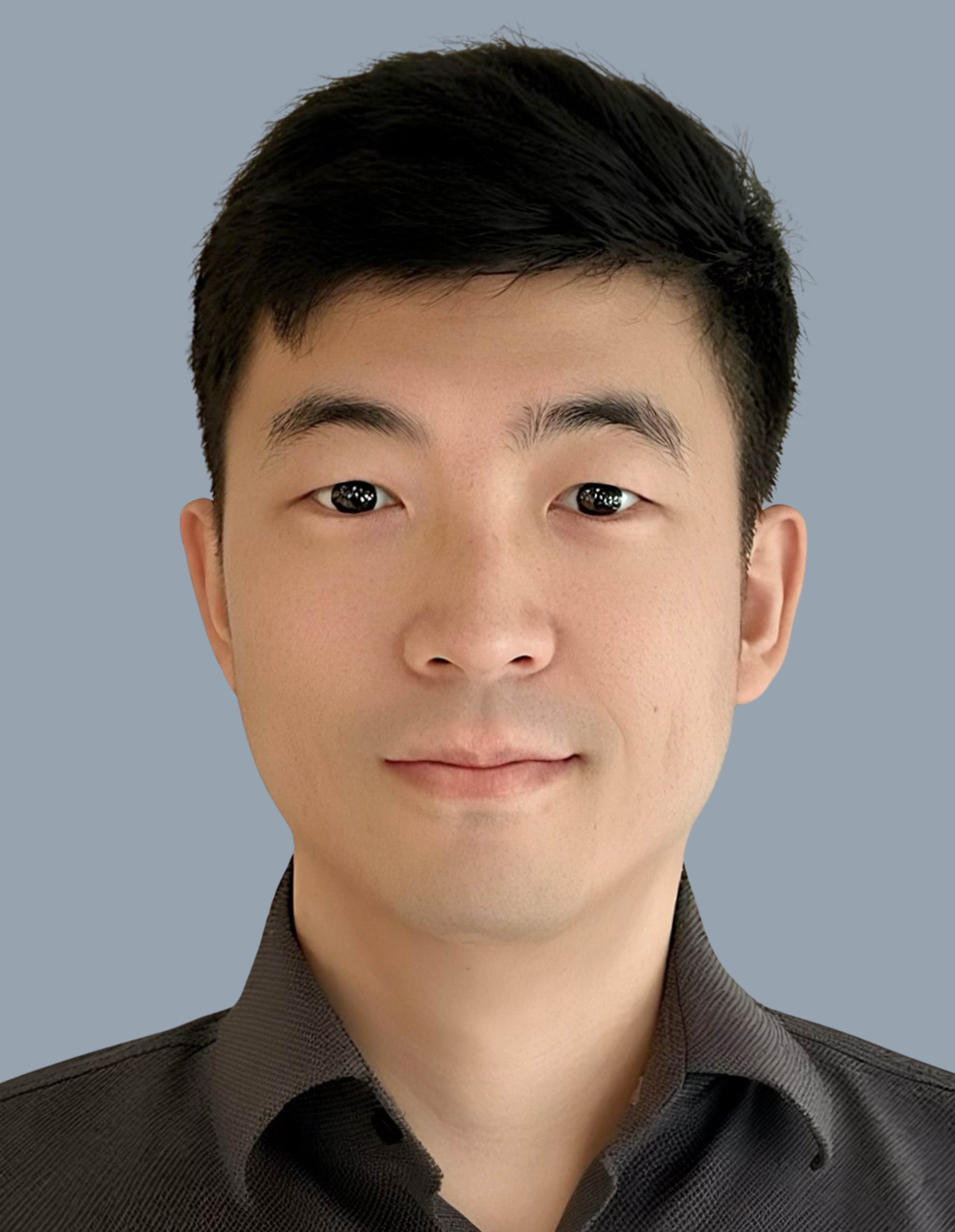}}]
{Yushuai Li} (Senior Member, IEEE) received the Ph.D. degree in control theory and  control engineering from the Northeastern University, Shenyang, China, in 2019.  He is currently an Assistant Professor with the Department of Computer Science, Aalborg University, Aalborg, Denmark.  He received the Best Paper Awards from Journal of Modern Power Systems and Clean Energy (MPCE) in 2021, and ICCSIE in 2023, IEEE EI2 in 2024. He serves as Associate Editors for IEEE Transactions on Industrial Informatics, IEEE Transactions on Automation Science and Engineering, MPCE, and IEEE Systems, Man, and Cybernetics Magazine. His main research interests include machine learning, digital twin, and integrated energy and transportation systems.
\end{IEEEbiography}

\vskip -1\baselineskip plus -1fil
\begin{IEEEbiography}[{\includegraphics[width=1in,height=1.25in,clip,keepaspectratio]{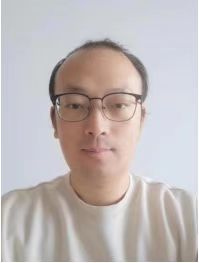}}]{Peiyuan Guan} received his M.Eng and Ph.D. degrees in software engineering and computer science technology from Central South University, Changsha, China, in 2016 and 2021, respectively. Currently, he is a researcher at the Department of Informatics, University of Oslo. He was a TPC member of the VTC 2020 fall. He was a reviewer for several journals, including IEEE TITS, IEEE TII, IEEE TCOM, IEEE IoT, etc. His research interests span several areas in communications and networking, with a focus on edge and distributed computing.
\end{IEEEbiography}

\vskip -1\baselineskip plus -1fil
\begin{IEEEbiography}[{\includegraphics[width=1in,height=1.25in,clip,keepaspectratio]{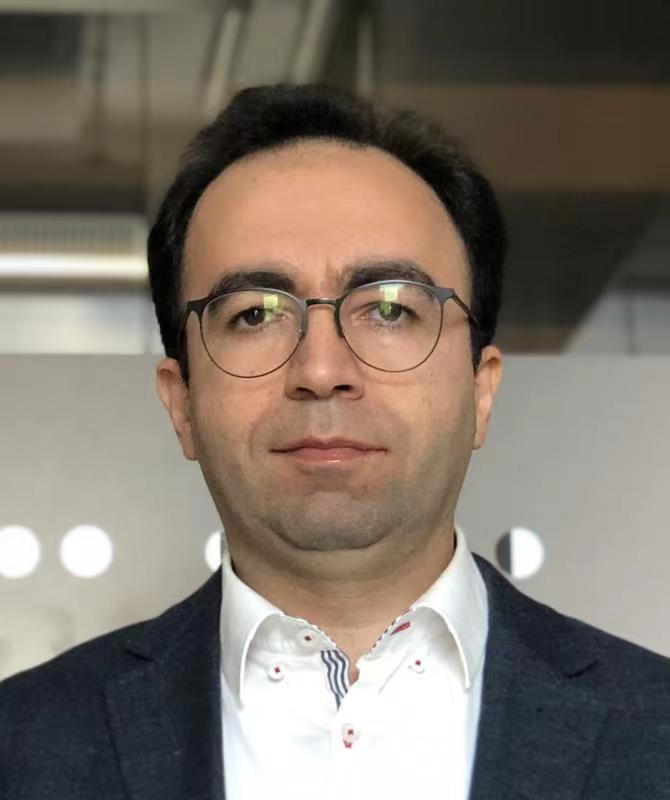}}]{Amir Taherkordi} received the Ph.D. degree in computer science from the University of Oslo, Oslo, Norway, in 2011. He is currently a Full Professor with the Department of Informatics, UiO. From 2013 to 2018, he was a Researcher with the Networks and Distributed Systems Group, Department of Informatics, UiO. He has experience from several national (Norwegian Research Council) and international (European
research funding agencies) research projects. He is an associate editor of IEEE Transactions on Network Science and Engineering and IEEE Transactions on Mobile Computing. His research interests include resource-efficiency, scalability, adaptability, dependability, mobility, and data-intensiveness of distributed systems designed for emerging computing technologies, such as the IoT, fog/edge/cloud computing, and cyberphysical systems.
\end{IEEEbiography}
\vskip -1\baselineskip plus -1fil
\begin{IEEEbiography}[{\includegraphics[width=1in,height=1.25in,clip,keepaspectratio]{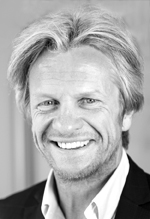}}]{Tor Skeie} is a professor with the University of Oslo and
Simula Research Laboratory; His research has mainly contributed to the High-Performance Computing (HPC) field. Herein he has focused on effective routing, fault tolerance, congestion control, quality of service, reservoir simulations, edge and cloud computing. He is also a researcher with the
Industrial Ethernet and wireless networking areas. Several of his research results have been published in the most respected IEEE Transactions and magazines. 
\end{IEEEbiography}

\end{document}